\documentclass{article}

\usepackage[preprint]{corl_2026} % Uncomment for pre-prints (e.g., arxiv); This is like ``final'', but will remove the CORL footnote.

\usepackage{acronym}
\usepackage{xspace}
\usepackage{amsmath}
\usepackage{amssymb}
\usepackage{comment}
\usepackage[table]{xcolor}
\usepackage{graphicx}
\usepackage{subcaption}
\usepackage{wrapfig}
\usepackage{array}
\usepackage{booktabs}
\usepackage{multirow}
\usepackage{tabularx}
\usepackage[most]{tcolorbox}

\title{FlowMaps: Modeling Long-Term Multimodal Object Dynamics with Flow Matching}

\author{
\textbf{Francesco Argenziano}\textsuperscript{1} \quad
\textbf{Miguel Saavedra-Ruiz}\textsuperscript{2,3} \quad
\textbf{Sacha Morin}\textsuperscript{2,3} \\
\textbf{Charlie Gauthier}\textsuperscript{2,3} \quad
\textbf{Daniele Nardi}\textsuperscript{1} \quad
\textbf{Liam Paull}\textsuperscript{2,3} \\[0.5em]
{\normalfont\textsuperscript{1}Sapienza University of Rome, Rome, Italy} \\
{\normalfont\textsuperscript{2}Université de Montréal, Montréal, QC, Canada} \\
{\normalfont\textsuperscript{3}Mila - Quebec AI Institute, Montréal, QC, Canada} \\[0.5em]
{\normalfont\texttt{\{argenziano,nardi\}@diag.uniroma1.it}} \\
{\normalfont\texttt{\{miguel-angel.saavedra-ruiz,sacha.morin\}@mila.quebec}} \\
{\normalfont\texttt{\{charlie.gauthier,paulll\}@mila.quebec}}
}

\begin{document}
\maketitle
\newcommand{\m}[1]{\mathcal{#1}}
\newcommand{\mb}[1]{\mathbb{#1}}
\newcommand{\mbf}[1]{\mathbf{#1}}

\newcommand{\fm}{\texttt{FlowMaps}\xspace}
\newcommand{\eg}{\emph{e.g.,}\xspace}
\newcommand{\ie}{\emph{i.e.,}\xspace}
\newcommand{\boldparagraph}[1]{\textbf{#1.}}

\definecolor{hab1}{RGB}{74,35,119}
\definecolor{hab2}{RGB}{89,168,156}
\definecolor{hab3}{RGB}{245,95,116}

\definecolor{lightgreenrow}{RGB}{238,249,242}
\definecolor{darkgreenrow}{RGB}{222,243,230}
\newcommand{\best}[1]{\cellcolor{darkgreenrow}\textbf{#1}}
\newcommand{\second}[1]{\cellcolor{lightgreenrow}#1}
\newcommand{\grey}[1]{\cellcolor{gray!20}#1}

\acrodef{FM}{flow matching}
\acrodef{VAE}{Variational Autoencoder}
\acrodef{DiT}{Diffusion Transformer}
\acrodef{CDiT}{Conditional Diffusion Transformer}
\acrodef{ObjNav}{Object Navigation}
\acrodef{ODE}{Ordinary Differential Equation}
\acrodef{CFM}{conditional flow matching}
\acrodef{MLP}{Multi Layer Perceptron}
\acrodef{LLM}{Large Language Model}
\acrodef{VLM}{Visual Language Model}
\acrodef{GNN}{Graph Neural Network}
\begin{abstract}
Joint spatial and temporal understanding of 3D scenes is a crucial requirement for robots deployed in everyday household environments. Such agents must not only comprehend and navigate spatial layouts, but also reason about how these spaces evolve over time. In particular, humans interact with objects daily, causing them to change position throughout the environment and making it difficult for robots to reliably associate current observations with previously seen objects. However, these interactions are not random: human habits and routines induce spatio-temporally consistent patterns in object locations, which robotic agents can potentially learn and then exploit for downstream tasks such as navigation. To this end, we introduce \fm, a latent flow matching model for estimating multimodal distributions over the future locations of dynamic objects in a continuous 3D space. By learning the implicit dependencies among objects and their temporal evolution, \fm predicts likely changes in object locations conditioned on past human interactions, while supporting generalization across previously unseen environments that share similar object routines. To demonstrate the utility of this method, we deploy \fm in a downstream dynamic Object Navigation task in both simulated and real-world environments. Across more than 600 episodes, \fm outperforms state-of-the-art approaches, showing that modeling object dynamics through continuous, multimodal spatio-temporal distributions improves robotic search and navigation in changing household environments. Code and additional material is available at https://fra-tsuna.github.io/flowmaps/.
\end{abstract}

\keywords{Flow matching, Dynamic environments, Object navigation} 

\section{Introduction}
\label{sec:1_intro}

Environments inhabited by humans are intrinsically dynamic. This dynamism should not be attributed solely to the people living in them, but also to the everyday objects they contain. Through repeated human interactions, objects are frequently moved, displaced, and relocated over time, posing major challenges for robots expected to deliberate and act in such environments~\cite{mois2020role, lopez2021low}. Under these conditions, even a seemingly simple task (\eg ``help me find my glasses''~\cite{10160473}) becomes highly challenging: although the robot may have observed an object in the past, there is no guarantee that it will still be found at its previous location, and reasoning about its possible displacement requires considering where it may have moved in physical space.
Humans constantly move objects within and across rooms, creating significant difficulties for navigation and retrieval. Nevertheless, humans tend to follow repetitive behavioral patterns~\cite{troje2008retrieving}. 
For instance, a pair of glasses may be placed on the nightstand before sleep, moved to the bathroom sink before a shower, and later returned to the nightstand. These routines are referred to as \emph{semantically consistent patterns}~\cite{schmid2022panoptic, yugay2025gaussian}, and can potentially be leveraged to recover from failed retrieval attempts by directing the robot toward likely human-induced object placements.

We hypothesize that, given sufficient observations of how objects move through environments over time, a generative model can learn latent regularities induced by human routines and exploit them to predict likely future object placements. Crucially, our key insight is that predicting where objects will be found does not require explicitly identifying the human activity that caused their displacement; instead, such structure can emerge directly from data. To be useful beyond a single observed home, these predictions should generalize to environments with different layouts and object arrangements, while remaining grounded in actual object locations rather than fixed, predefined alternatives such as receptacle labels. Since object displacements are inherently multimodal (\ie they admit multiple \emph{modes}, each corresponding to a distinct plausible placement), modeling them requires a distributional approach over continuous space.

To this end, we present \fm, a latent \ac{FM} model that recovers multimodal spatio-temporal distributions of common household objects directly in continuous 3D space. Unlike prior dynamic object location models that primarily operate over discrete receptacle-level relations, \fm models human-induced household object relocalization as a multimodal distribution in continuous 3D space. \fm is composed of two main modules: \emph{(i)} a \ac{VAE} that learns latent representations of object geometry and semantics, and \emph{(ii)} a latent \ac{DiT}~\cite{peebles2023scalable} that predicts likely object locations over time.
To obtain training data at scale, we employ ProcTHOR~\cite{procthor} to generate meaningful dynamic object trajectories across procedurally generated household environments.
We then demonstrate the usefulness of the learned prior on \ac{ObjNav}, using it as a representative downstream robotic task among the broader set of applications that can benefit from predictions of likely object placements.
In this setting, \fm is trained on dynamic trajectories from a set of training environments, while \ac{ObjNav} performance is evaluated in disjoint, previously unseen homes, explicitly testing whether the learned prior transfers beyond the scenes observed during training.
Across more than 600 \ac{ObjNav} episodes in ProcTHOR, we compare \fm against state-of-the-art approaches and show superior performance in retrieving target objects.
We further validate our approach on a real robotic platform, showing how it can be deployed in a real-world setting.

\boldparagraph{Contributions statement} Our contributions are three-fold: \textbf{\emph{(i)}} \fm, a \ac{FM}-based architecture for representing and predicting continuous, multimodal spatio-temporal object distributions in dynamic environments; \textbf{\emph{(ii)}} as an exemplary downstream robotic application, we demonstrate the use of \fm for \ac{ObjNav} in dynamic scenes, enabling a robot to reason about where target objects are likely to be found over time, while outperforming both zero-shot baselines and trained expert policies;
\textbf{\emph{(iii)}} an extensive quantitative and qualitative evaluation of both the distributional properties and practical applicability of \fm, covering more than 600 simulated \ac{ObjNav} episodes in disjoint environments together with real-world deployments, and demonstrating the soundness, effectiveness, cross-environment generalization, and practical viability of the proposed approach.

%The remainder of this paper is organized as follows. Section~\ref{sec:2_rw} reviews related work, while Section~\ref{sec:3_bg} introduces the background required for the \ac{FM} framework and formalizes the problem addressed in this work. Section~\ref{sec:4_fm} then presents \fm, followed by the experimental results in Section~\ref{sec:5_results}. Section~\ref{sec:6_lfd} highlights current limitations and possible future directions, and Section~\ref{sec:7_conc} concludes the paper.
\section{Related Work}
\label{sec:2_rw}

\boldparagraph{Learning human habits and patterns}
Learning human habits enables robots to move from reactive command execution to proactive assistance~\cite{li2015personalizing,irfan2019personalization}. Prior work models human activities for anticipation~\cite{sung2012unstructured,koppula2015anticipating}, uses interaction histories for personalized collaboration~\cite{baraglia2017efficient},
and studies anticipatory robot assistance~\cite{hoffman2007effects,buyukgoz2022two,van2024proactive}. The closest work to ours is HOMER~\cite{pmlr-v205-patel23a}, which uses a \ac{GNN} to predict object displacements from recurring household activities.
Unlike HOMER, which trains a separate model per household and mainly evaluates within-environment generalization, we train and test on disjoint environments sharing semantically consistent patterns, assessing generalization to unseen household layouts.

\boldparagraph{Object navigation in dynamic environments}
\ac{ObjNav} aims to locate target objects in known or unknown environments. Existing methods often rely on \acp{LLM} or \acp{VLM} for zero-shot reasoning~\cite{dorbala2023can,rajvanshi2024saynav}, or learn end-to-end policies~\cite{zhu2017target}, but typically assume static scenes. Recent dynamic approaches incorporate object trajectories~\cite{dorbala2026personalizedembodiednavigationportable,dorbala2023can}, probabilistic object-location estimates~\cite{10160473,wang2024dynamic}, or dynamic scene graph memories~\cite{kurenkov2023modeling}, yet they depend on costly \ac{VLM} reasoning, hand-designed priors, online estimation, \ac{LLM}-interpreted activity hints, or explicit graph structures, and have often been evaluated in limited settings. In contrast, \fm models object dynamics as continuous distributions over future object placements in 3D, rather than discrete temporal link prediction over object-location relations~\cite{pmlr-v205-patel23a, kurenkov2023modeling}. It requires no explicit scene graph structure at training or inference time, remains efficient through its \ac{FM} formulation, and outperforms prior dynamic \ac{ObjNav} methods.

\boldparagraph{Flow matching in robotics}
Generative models are increasingly used in robotics, supported by large cross-embodiment datasets such as Open-X Embodiment~\cite{o2024open}. Recent work uses diffusion models~\cite{chi2024diffusionpolicy,hou2024diffusion}, \ac{FM}~\cite{chisari2025learning,zhang2024affordance}, and vision-language-action models~\cite{black2024pi0,kim2025openvla} mainly for policy learning. In contrast, we use \ac{FM} for posterior inference, recovering spatio-temporal and multimodal distributions over plausible object placements rather than generating actions. To the best of our knowledge, this is the first use of \ac{FM} for posterior inference in this setting.

\section{Background and problem formulation}
\label{sec:3_bg}
\boldparagraph{Vector fields and flows}
An \ac{ODE} is defined by a time-dependent \emph{vector field}
$u: \mathbb{R}^d \times [0,1] \rightarrow \mathbb{R}^d$, which assigns a velocity
$u_t(x)$ to every position $x \in \mathbb{R}^d$ at each time $t \in [0,1]$. A solution of the \ac{ODE} is a \emph{trajectory} $X:[0,1] \rightarrow \mathbb{R}^d$, where $X_t$ denotes the position of the system at time $t$. In this work, we are particularly interested in \emph{flows} as solutions of the \ac{ODE}. A flow is a time-dependent map $\psi:\mathbb{R}^d \times [0,1] \rightarrow \mathbb{R}^d$, written
$\psi_t(x) = \psi(x,t)$, whose evolution is governed by
$$
\frac{d}{dt}\psi_t(x) = u_t(\psi_t(x)),
    \qquad
    \psi_0(x_0) = x_0 .
$$

The vector field $u_t$ is said to generate a \emph{probability path}
$(p_t)_{0 \leq t \leq 1}$ if the corresponding flow transports samples from
the initial distribution $p_0$ to the distribution $p_t$ at each time $t$.
Equivalently, for $X_0 \sim p_0$, we have $X_t = \psi_t(X_0) \sim p_t$ .

\boldparagraph{Conditional flow matching} \Ac{CFM} learns such a vector field by regressing onto tractable conditional velocities. The goal is to learn $u_t^\theta$ whose flow transports a simple base distribution $p_{init}$ to the data distribution $p_{data}$. Instead of directly constructing the marginal vector field, namely the vector field that generates the full probability path from $p_{init}$ to $p_{data}$, \ac{CFM} defines conditional paths $p_t(\cdot \mid y)$ indexed by data samples $y\sim p_{data}$, together with tractable target velocities $u_t^{target}(\cdot \mid y)$. The training objective is
\begin{equation}
\label{eq:cfm}
{\small
\mathcal{L}_{CFM}(\theta)
=
\mathbb{E}_{t,x,y}
\!\left[
\left\|u_t^\theta(x)-u_t^{\mathrm{target}}(x\mid y)\right\|_2^2
\right],
\quad
t\sim U[0,1],\ y\sim p_{data},\ x\sim p_t(\cdot\mid y).
}
\end{equation}
Under standard assumptions, this objective has the same gradient as the corresponding marginal \ac{FM} objective, so minimizing \eqref{eq:cfm} learns the marginal vector field without requiring its explicit evaluation. For a more complete explanation and formalization, we refer the reader to~\cite{lipman2024flowmatchingguidecode,flowsanddiffusions2026}.

\boldparagraph{Latent flow matching}
In \emph{latent flow matching}, the same learning objective is applied in a
latent representation space. Given an encoder $E$ and decoder $D$, data samples $x \sim p_{data}$ are mapped to latents $z=E(x)$, inducing a latent data distribution $p_{data}^{z}$. A vector field $u_t^\theta$ is then trained to transport samples from a simple latent prior $p_{init}^{z}$ to $p_{data}^{z}$ by minimizing the \ac{CFM} objective in latent space. Sampling draws $Z_0 \sim p_{init}^{z}$, integrates $\frac{d}{dt}Z_t = u_t^\theta(Z_t)$, and decodes the terminal latent variable as $\hat{x}=D(Z_1)$. This reduces the dimensionality of the transport problem and
lets the flow operate on compact, semantically structured representations, while the decoder maps generated latents back to the data domain.

\boldparagraph{Problem formulation}
At time \(\tau \geq 0\), we represent the map as
\(M_\tau = (O_\tau, O_{\mathrm{BG}})\), where
\(O_\tau = \{O_{i,\tau}\}_{i=1}^{N_O}\) denotes the set of dynamic objects
observed in the scene, and \(O_{\mathrm{BG}}\) denotes the static background,
which is assumed to remain unchanged over time. Each dynamic object is
represented as
$O_{i,\tau} = (\mathbf{b}_{i,\tau}, l_i)$,
where $\mathbf{b}_{i,\tau} \in \mathbb{R}^6$ denotes the object's 3D axis-aligned
bounding box in center-size format, $\mathbf{b}_{i,\tau} = (c_{i,\tau}^x, c_{i,\tau}^y, c_{i,\tau}^z, s_i^x, s_i^y, s_i^z)$, with $(c_{i,\tau}^x, c_{i,\tau}^y, c_{i,\tau}^z)$ denoting the box center at time $\tau$, and $(s_i^x, s_i^y, s_i^z)$ denoting its spatial extent. The label $l_i \in \mathcal{L}$ denotes the semantic class of the object. In each scene, there is at most one dynamic object per semantic class, so no two dynamic objects share the same label. Background elements in $O_{\mathrm{BG}}$ are represented analogously, but their spatial states are fixed across time.
Given a prediction horizon \(\Delta \tau \geq 0\) and an object query label
%text-object query
\(l_q\) identifying a dynamic object in \(O_\tau\), our goal is to infer the distribution over future bounding boxes of
the queried object at time \(\tau_f = \tau + \Delta \tau\). Namely, we want to infer the distribution
$p\left(\mathbf{b}_{q,\tau_f} \mid M_\tau, l_q, \tau_f \right)$.
\begin{comment}    
Directly computing this posterior is generally intractable. We therefore
employ \ac{FM} to approximate it and to generate samples from it.
In particular, starting from a random sample~\(z_{\mathrm{init}} \sim p_0\) drawn from a simple base distribution \(p_0 = \mathcal{N}(0,I)\), we learn a conditional flow that transports \(p_0\) to the target distribution
$$ p_1 = p\left(\mathbf{b}_{q,\tau_f} \mid M_\tau, l_q, \tau_f \right). $$
After integrating the learned flow, the resulting sample $z_{\mathrm{final}} \sim p_1$ represents a plausible future bounding box of the queried object at time $\tau_f$.
\end{comment}

\section{FlowMaps}
\label{sec:4_fm}
Directly computing this posterior is intractable, as the queried object's future
state depends on rich scene context and may admit multiple plausible outcomes.
We approximate it with \ac{FM}, learning a conditional transport from a Gaussian
base distribution \(p_0 = \mathcal{N}(0,I)\) to the target distribution
\[
p_1 = p\left(\mathbf{b}_{q,\tau_f} \mid M_\tau, l_q, \tau_f \right).
\]
Starting from \(z_{\mathrm{init}} \sim p_0\), integrating the learned flow
produces a sample \(z_{\mathrm{final}} \sim p_1\), corresponding to a plausible
future bounding box of the queried object at time \(\tau_f\).
\begin{comment}
    
Directly computing the aforementioned posterior is intractable, as the future state of the queried object is conditioned on rich scene context and may admit
multiple plausible outcomes. We therefore approximate it using \ac{FM}, which enables sampling from the target distribution by learning a continuous transport from a simple base distribution. In particular, starting from a random sample \(z_{\mathrm{init}} \sim p_0\), with \(p_0 = \mathcal{N}(0,I)\), we learn a conditional flow that maps \(p_0\) to
$
p_1 = p\left(\mathbf{b}_{q,\tau_f} \mid M_\tau, l_q, \tau_f \right).
$
After integrating the learned flow, the resulting sample \(z_{\mathrm{final}} \sim p_1\) represents a plausible future bounding box of the queried object at time \(\tau_f\).
\end{comment}

We instantiate this conditional flow in latent space with \fm, a Transformer-based \ac{FM} model built on the \ac{CDiT} of~\citet{bar2025navigation}. A \ac{VAE} first encodes object bounding boxes and semantic labels into a latent target space (Section~\ref{sec:4.2:vae}). A \ac{CFM} network then learns the flow in this space, predicting future object placements conditioned on the current scene, queried label, and prediction horizon (Section~\ref{sec:4.3:fm-network}). Dynamic scene generation and data collection are described in Section~\ref{sec:4.1:pre}. Architectural, hyperparameter, solver, and interpolation details are provided in the supplementary material.

\subsection{Dynamic Scenes Generation and Data Collection}
\label{sec:4.1:pre}

\begin{comment}
    
\begin{wrapfigure}[15]{r}{0.6\linewidth}
    \centering
    \vspace{-1.0em}
    \includegraphics[width=\linewidth]{figures/habits_2weeks.pdf}
    \caption{Different spatio-temporal behaviors of the same object (``Book") under different human habits in a ProcTHOR scene. Patterns displayed for 2 weeks.}
    \label{fig:habits}
\end{wrapfigure}
\end{comment}
We used ProcTHOR~\cite{procthor} to generate dynamic indoor environments for training both components of \fm. Object movements are driven by predefined human-like routines that produce semantically consistent patterns. We instantiated three representative routines, each capturing a different type of plausible indoor behavior. In~\textcolor{hab1}{Habit~\#1}, the simulated human exhibits location preferences, repeatedly returning to a small set of favored places and therefore spending more time there. In \textcolor{hab2}{Habit~\#2}, the simulated human follows a balanced routine, distributing time approximately uniformly across the relevant locations in the environment. In \textcolor{hab3}{Habit~\#3}, the simulated human follows a highly dynamic routine, frequently transitioning between locations and spending only short intervals at each one. For each routine, we generate 2706 training and 918 validation environments. Each scene contains up to 15 dynamic objects moving between semantically compatible receptacles, \eg a ``Fork'' can appear on a ``Sink'' or ``DiningTable'', but not on a ``ShelvingUnit''. Each environment is simulated for 4 weeks with hourly resolution $d\tau$, yielding 672 timestamps per scene $s$, with $s_{\tau} = [\tau, \m{O}{\tau}, \m{O}{BG}]$. This results in over 1.8M training samples per habit. For each habit, we train and evaluate a separate \fm model. More information on habits can be found in the appendix.

\subsection{Variational Autoencoder}
\label{sec:4.2:vae}

\begin{comment}
\begin{wrapfigure}[12]{r}{0.5\linewidth}
    \centering
    \vspace{-0.7em}
    \includegraphics[width=\linewidth]{figures/vae.pdf}
    \caption{\ac{VAE} used to encode and decode object tokens in a ProcTHOR environment. Dotted bounding boxes mean decoded objects.}
    \label{fig:vae}
    %\vspace{-1em}
\end{wrapfigure}
\end{comment}
We train a \ac{VAE} to map individual object tokens into the latent space. Each token consists of a normalized 3D bounding box $\mathbf{b}$ and a semantic label $l$, and is encoded into a latent code $\mathbf{z} \in \mathbb{R}^{d_z}$. Once trained, the \ac{VAE} is frozen: its encoder $e_\phi$ provides the target latents for the \ac{CDiT}, while its decoder $d_\xi$ maps generated latents back to object predictions. The decoder is trained to predict both geometry and semantics: its geometry head reconstructs the normalized bounding box parameters $\hat{\mathbf{b}}$, while its semantic head predicts logits over object classes. The network is trained with the standard \ac{VAE} objective 
$ \mathcal{L}_\text{VAE} =
  \mathcal{L}_\text{rec}
  +
  \beta_t \, \mathcal{L}_\text{KL},
$
where $\beta_t$ is linearly annealed during training to prevent posterior collapse~\cite{bowman2016generating}. The reconstruction term combines geometric and semantic supervision:
$
  \mathcal{L}_\text{rec}
  =
  \lambda_\text{CIoU}\,\mathcal{L}_\text{CIoU}
  +
  \lambda_{L_1}\,\mathcal{L}_{L_1}
  +
  \lambda_\text{CE}\,\mathcal{L}_\text{CE}.
$
Here, $\mathcal{L}_\text{CIoU}$ and $\mathcal{L}_{L_1}$ supervise the reconstructed bounding box, while $\mathcal{L}_\text{CE}$ supervises the semantic class prediction.

\subsection{Latent Flow Matching network}
\label{sec:4.3:fm-network}
The \fm network parametrizes the velocity field $u^{\theta}_t$ that transports a Gaussian latent $\mathbf{z}_0$ to the latent encoding $\mathbf{z}_1 = e_\phi(\mathbf{b}_q, l_q)$ of the queried object's future bounding box. 
It is composed of: \emph{(i)} a \emph{map encoder} that aggregates the scene context into a sequence of contextualized tokens, and \emph{(ii)} a stack of \emph{\ac{CDiT} blocks} that iteratively refine the noisy query latent by cross-attending to that context.%; both are observable in Fig.~\ref{fig:flowmaps_architecture}.

\begin{wrapfigure}[21]{r}{0.40\columnwidth}
    \centering
    \begin{subfigure}{\linewidth}
        \centering
        \vspace{-1.3em}
        \includegraphics[width=\linewidth]{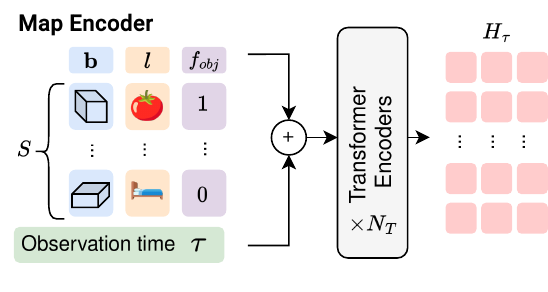}
        \vspace{-1.6em}
        \caption{Map encoder.}
        \label{fig:map_encoder}
    \end{subfigure}

    \begin{subfigure}{\linewidth}
        \centering
        \includegraphics[width=\linewidth]{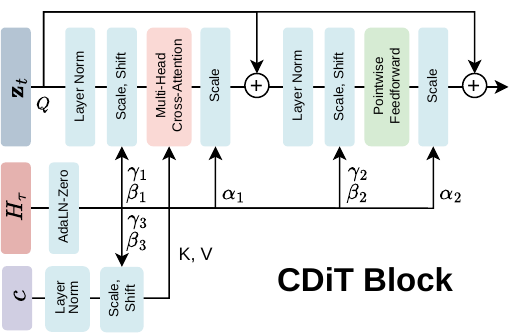}
        \caption{\ac{CDiT} block.}
        \label{fig:cdit_block}
    \end{subfigure}

    \caption{Overview of the latent \ac{FM} network components: \emph{(a)} the map encoder, and \emph{(b)} the \ac{CDiT} block.}
    
    \label{fig:flowmaps_architecture}
\end{wrapfigure}

\begin{comment}
\boldparagraph{Map encoder} 
At each timestep, the scene is represented as a padded set of at most $N_O + N_{BG}$ object tokens composed, with maximum sequence length $S$. Each token corresponds to either a dynamic object or a static background object, and consists of a 3D axis-aligned bounding box $\mathbf{b}$, a semantic class label $l$, and an object-type flag $f_{obj}$. Following the general idea of appending set-membership indicators from \citet{wald2020learning}, we use $f_{obj}$ to distinguish the two object sets in our representation: dynamic objects ($f_{obj}=1$) and furniture/background objects ($f_{obj}=0$).  The map encoder (Fig.~\ref{fig:map_encoder}) turns the tokenized scene $\mathcal{M}_\tau$ into a sequence of $S$ context tokens $H_\tau \in \mathbb{R}^{S \times d_h}$. The per-token embeddings of bounding box, semantic class, and type flag are summed together with a learned embedding of the observation time $\tau$, which anchors the scene to its temporal slice.  These tokens are processed by a stack of $N_T$ pre-norm Transformer encoder layers with self-attention, allowing each object representation to incorporate scene-level context and relations to furniture and to the other dynamic objects. We deliberately keep the encoder permutation-invariant (\ie \emph{no positional encoding} is added across tokens): we treat the scene as a \emph{set} of objects rather than an ordered sequence, and spatial structure is conveyed only through the bounding box embedding.
\end{comment}

\boldparagraph{Map encoder}
At each timestep, the scene is a padded set $\mathcal{M}_\tau$ of at most $N_O+N_{BG}$ object tokens, with maximum length $S$. Each token represents either a dynamic object or a static background object through a 3D axis-aligned bounding box $\mathbf{b}$, semantic label $l$, and object-type flag $f_{obj}$. Following \citet{wald2020learning}, this flag acts as a set-membership indicator, where $f_{obj}=1$ for dynamic objects and $f_{obj}=0$ for furniture/background objects. The map encoder (Fig.~\ref{fig:map_encoder}) maps $\mathcal{M}_\tau$ to context tokens $H_\tau \in \mathbb{R}^{S \times d_h}$ by summing bounding box, class, type, and learned time embeddings, then processing them with $N_T$ pre-norm Transformer encoder layers. Self-attention lets each token incorporate scene-level context and relations to other objects. No positional encoding is used across tokens, making the encoder permutation-invariant: spatial structure is provided only by the bounding box embedding.

\boldparagraph{\ac{CDiT} block} Given $H_\tau$, the flow is implemented by $N_C$ \ac{CDiT} blocks (Fig.~\ref{fig:cdit_block}) that update the noisy query latent
$\mathbf{z}_t \in \mathbb{R}^{d_z}$ by cross-attending to the encoded scene. This \ac{CDiT}-style design, rather than full-attention \ac{DiT}
\cite{peebles2023scalable}, reflects the asymmetry of the task: a single query object is transported while the scene provides a variable-length context,
avoiding repeated quadratic attention over the full scene at each \ac{FM} step. Cross-attention and feed-forward layers are modulated with
\emph{adaLN-Zero}~\cite{peebles2023scalable} using a conditioning vector $c$ built from the flow time $t \in [0,1]$, final timestamp $\tau_f$, and queried
label $l_q$. Since the residual stream contains only one query token, we remove the per-block self-attention of~\citet{bar2025navigation}; scene-level reasoning
is handled by the map encoder, while the \ac{CDiT} blocks condition the query trajectory through cross-attention. A final adaLN-modulated linear head maps the
refined token to the \ac{VAE} latent space, producing $u^{\theta}_t(\mathbf{z}_t \mid H_\tau, l_q, \tau_f)$.

\begin{figure}[t!]
    \centering
    \includegraphics[width=\linewidth]{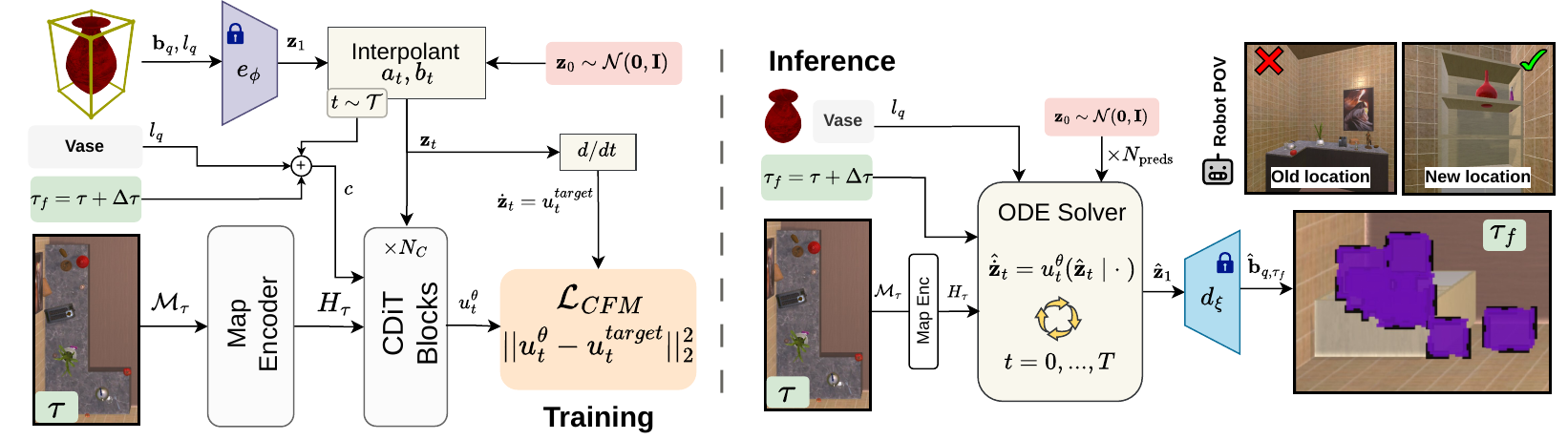}
    \caption{Training (left) and inference (right) pipeline for \fm. We train the stack of \ac{CDiT} blocks to learn the vector field $u_t^\theta$, which is used at inference time to predict $\hat{\mathbf{b}}_{q,\tau_f}$.}
    \label{fig:pipeline}
\vspace{-1.5em}
\end{figure}

\boldparagraph{Training and inference} We train the network with the \ac{CFM} loss of Eq.~\ref{eq:cfm} on a conditional optimal-transport path~\cite{lipmanflow}. Target latents $\mathbf{z}_1$ are obtained by encoding ground-truth future bounding boxes with the frozen \ac{VAE} encoder $e_\phi$ and standardizing them with training-set statistics, while source latents $\mathbf{z}_0 \sim \mathcal{N}(\mathbf{0}, \mathbf{I}_{d_z})$ are sampled independently. For $t \sim \mathcal{T}$ on $[0,1]$, we set $\mathbf{z}_t = a_t\mathbf{z}_1 + b_t\mathbf{z}_0$ and regress $u_t^\theta$ to $\dot{\mathbf{z}}_t=\dot{a}_t\mathbf{z}_1+\dot{b}_t\mathbf{z}_0$ with an $L_2$ loss. Source-target pairs are matched by an exact mini-batch optimal-transport plan~\cite{tong2024improving, pmlr-v202-pooladian23a}, yielding straighter trajectories and fewer integration steps. At inference, given $\mathcal{M}_\tau$ and $(l_q,\tau_f)$, one map-encoder pass computes the query-independent $H_\tau$. Since $H_\tau$ is query-independent, the same encoder pass is amortized across all dynamic objects and posterior samples. We then integrate $\hat{\dot{\mathbf{z}}}_t=u_t^\theta(\hat{\mathbf{z}}_t\mid H_\tau,l_q,\tau_f)$ from $\mathbf{z}_0$ to $\hat{\mathbf{z}}_1$ in $K$ fixed steps, de-standardize $\hat{\mathbf{z}}_1$, and decode it with $d_\xi$ into $\hat{\mathbf{b}}_{q,\tau_f}$. Independent $\mathbf{z}_0$ draws produce samples from $p(\mathbf{b}_{q,\tau_f}\mid \mathcal{M}_\tau,l_q,\tau_f)$. The full pipelines are shown in Fig.~\ref{fig:pipeline}.

\section{Experimental Results}
\label{sec:5_results}

\boldparagraph{Distributional Evaluation}
We evaluate \fm using the joint accuracy and distributional metrics reported in Table~\ref{tab:eval_metrics}. All entries are averaged over the three considered habits.

We report two best-of-$K$ accuracy metrics, following the protocol introduced for multimodal trajectory prediction~\cite{gupta2018socialgan}. For each query, we draw $K{=}50$ samples. \emph{minFDE@K} measures the 3D distance between the ground-truth bounding box center and the closest predicted sample, while \emph{Rec@1} reports the corresponding best-of-$K$ recall within $\delta{=}1$\emph{m} on the floor plane. To assess the predicted distribution beyond point accuracy, we also report \emph{coverage} and \emph{density} following \citet{pmlr-v119-naeem20a}. Coverage is the fraction of ground-truth positions whose $k$-Nearest Neighbor ball contains at least one generated sample, whereas density is the average number of such balls occupied by each sample. These metrics separate mode coverage from over-concentration, without penalizing valid multimodal predictions. We further compute \emph{TV} and \emph{JS}, respectively the total-variation and Jensen-Shannon divergences between predicted and ground-truth samples discretized as histograms on an xz floor grid with $0.5$\,m resolution. These divergences measure how spatial mass is allocated across the scene, independently of best-of-$K$ accuracy.

We compare \fm against four baselines. \textbf{FreqPrior} estimates a training-set distribution \(P(r \mid l_q)\) over receptacle categories \(r\). At test time, given \((\mathcal{M}_\tau,l_q)\), it samples a receptacle category, selects an instance of that category in the scene, and samples a point uniformly from the footprint of its top face. This captures scene-conditional object co-occurrence but ignores temporal structure. \textbf{LLMPrior} replaces this empirical distribution with a ranked top-\(T\) list of receptacles suggested by an LLM. \textbf{EmpiricalMean} is an oracle point predictor that returns the mean of all ground-truth positions of objects with label \(l_q\) in the environment, providing a strong upper bound for regression methods with the same \((\mathcal{M}_\tau,l_q)\) conditioning. \textbf{MeanFlowMaps} collapses \fm samples to their centroid, isolating the effect of multimodality from the conditioning signal. More details on the distributional baselines are provided in the appendix.

\fm achieves the best minFDE ($0.342$ \emph{m}, $7.6$\% below the strongest baseline), while nearly saturating mode coverage. It also raises the Naeem density from $0.647$ (FreqPrior) to $0.886$, a $37$\% gain, indicating that its samples concentrate more tightly around ground-truth modes. The two single-point baselines are consistently worse on accuracy (minFDE $>2$ \emph{m}, Rec@1 $<0.42$); MeanFlowMaps, despite using the same conditioning, is over $7{\times}$ worse in minFDE, confirming the need for a multimodal generator. FreqPrior matches \fm only in best-of-$K$ recall, whereas \fm also reduces the error tail, increases density, and lowers both TV and JS, showing that it captures \emph{where} on the plausible surfaces mass concentrates, not just \emph{which} surfaces are plausible.
%the distributional shape rather than only the correct region.

\begin{table}[t]
\centering
\scriptsize
\caption{Distributional evaluation averaged across the three habits. Density, coverage, TV and JS are distribution-shape metrics and are reported only for methods that emit a non-degenerate sample distribution.\looseness-1}
\label{tab:eval_metrics}
\begin{tabular}{lcccccc}
\toprule
Method & minFDE(\textit{m})$\downarrow$ & Rec@1$\uparrow$ & Density$\uparrow$ & Coverage$\uparrow$ & TV$\downarrow$ & JS$\downarrow$ \\
\midrule
FreqPrior        & \second{0.370} & \best{0.958}   & 0.647          & \second{0.991} & 0.848          & 0.514          \\
LLMPrior         & 0.907          & 0.780          & \second{0.773} & 0.967          & \best{0.796}   & \best{0.471}   \\
EmpiricalMean    & 2.040          & 0.414          & ---            & ---            & ---            & ---            \\
MeanFlowMaps     & 2.712          & 0.260          & ---            & ---            & ---            & ---            \\
\fm (ours) & \best{0.342} & \second{0.942} & \best{0.886} & \best{0.999} & \second{0.829} & \second{0.482} \\
\bottomrule
\end{tabular}
\vspace{-2.2em}
\end{table}
\boldparagraph{Object Navigation} We further evaluate \fm on \ac{ObjNav}, our primary downstream application, using a minival split of 25 random validation environments.
For each environment, we select 5 query objects and sample 5 $(\tau, \tau_f)$ pairs per object, obtaining more than 600 \ac{ObjNav} episodes.
An episode is successful if the agent reaches the target within $N_\text{steps}$ steps, is within distance $\delta_{\text{dist}}$ of the object's ground-truth position, and observes the object from its point of view.
All simulated experiments are conducted in AI2-THOR~\cite{ai2thor} with ProcTHOR~\cite{procthor} environments, where visibility is assessed using ground-truth information.
In the real-world lab demo, a \ac{VLM} is used only to trigger the candidate ``\emph{found}'' signal, while success is still evaluated using the same three criteria, so false positives do not count as successful episodes.

We compare \fm against representative dynamic \ac{ObjNav} baselines: \textbf{TAP-LGX}~\cite{dorbala2026personalizedembodiednavigationportable}, a zero-shot \ac{VLM}-based extension of LGX~\cite{dorbala2023can}; \textbf{OSG}~\cite{10160473} and \textbf{CEG}~\cite{wang2024dynamic}, which implement the CP-SAT planning approximation of~\citet{10160473}; \textbf{HOMER}~\cite{pmlr-v205-patel23a} and \textbf{SGM}~\cite{kurenkov2023modeling}, which model dynamic \ac{ObjNav} as temporal link prediction with \acp{GNN}, with SGM additionally exploiting the scene graph hierarchy; and a \textbf{Naive \ac{LLM}} baseline directly prompted to rank future object locations.
Since OSG and CEG require object location likelihoods, we follow~\citet{wang2024dynamic} and evaluate scene-prior (SP) variants based on object-receptacle co-occurrences, as well as \ac{LLM}-based variants that parse human habit hints into relevant events for posterior estimation.
All methods return a ranked list of $K$ candidate future locations; for \fm, we obtain this list by running inference $N_{\text{preds}}$ times, clustering predictions with DBSCAN~\cite{dbscan}, and ranking clusters by mass.
Additional baseline details are provided in the appendix.

We report Success Rate (SR) and Success weighted by Path Length (SPL)~\cite{anderson2018evaluation}, including SR@K, SPL@K, mean SR (mSR), and mean SPL (mSPL) averaged over $K$.
Success at smaller $K$ indicates that the model localizes the object's future position with fewer trials.
We also report the mean path length and mean number of steps required to reach the target.
Table~\ref{tab:objvav} shows that \fm consistently achieves the best mSR and mSPL across all settings, while remaining optimal or competitive on the other metrics.
Notably, \fm' SR@1 exceeds the SR@5 of some baselines, indicating that its first prediction can be more accurate than several competing top-5 proposal sets.
Among the baselines, \ac{GNN}-based methods perform best on the easier \textcolor{hab1}{Habit~\#1} and \textcolor{hab2}{Habit~\#2} settings, but are surpassed by zero-shot \ac{LLM}-based methods on the harder \textcolor{hab3}{Habit~\#3}.
This suggests that harder behavioral patterns benefit more from transferable dynamic exploration than from directly generalizing learned patterns.
Overall, \fm performs best on all three habits, showing stronger dynamic-pattern learning and cross-environment generalization.

\newcommand{\pmval}{xx.x$\pm$x.x}

\begin{table}[t!]
\label{tab:results}
\centering
\scriptsize
\setlength{\tabcolsep}{2.4pt}
\renewcommand{\arraystretch}{1.05}

\resizebox{\linewidth}{!}{%
\begin{tabular}{@{}llcccccccccc@{}}
\toprule
&
Method &
SR@1 (\%)$\uparrow$ &
SR@5 (\%)$\uparrow$ &
SR@10 (\%)$\uparrow$&
mSR (\%)$\uparrow$&
SPL@1 (\%)$\uparrow$&
SPL@5 (\%)$\uparrow$&
SPL@10 (\%)$\uparrow$&
mSPL (\%)$\uparrow$&
Path (\textit{m})$\downarrow$&
Steps$\downarrow$
\\
\midrule

%------- Gaussian -------

\multirow{9}{*}{\rotatebox{90}{\textcolor{hab1}{\textbf{Habit \#1}}}}
& Naive \ac{LLM}
& 42.13 & 59.34 & 63.77 & 57.44 & 33.54 & 38.40 & 39.03 & 37.80 & 12.01 & 195.1
\\

& TAP-LGX~\cite{dorbala2026personalizedembodiednavigationportable}
& 35.25 & 57.05 & 62.62 & 54.03 & 28.88 & 40.03 & 44.00 & 43.62 & 10.62 & 193.6
\\

& OSG+SP~\cite{10160473}
& 18.69 & 56.89 & 64.10 & 51.85 & 15.34 & 31.46 & 33.21 & 29.29 & 12.99 & 224.3
\\

& OSG+LLM~\cite{10160473}
& 42.79 & 62.46 & 63.44 & 60.13 & 34.65 & 44.03 & 44.27 & 42.90 & \second{8.38} & \second{131.9}
\\

& CEG+SP~\cite{wang2024dynamic}
& 22.30 & 58.03 & 64.75 & 52.66 & 18.16 & 33.32 & 34.82 & 30.98 & 12.42 & 212.7
\\

& CEG+LLM~\cite{wang2024dynamic}
& 42.95 & 62.62 & 63.61 & 60.29 & 34.93 & 44.57 & 44.81 & 43.40 & \best{8.00} & \best{126.3}
\\

& SGM~\cite{kurenkov2023modeling}
& \second{48.36} & \second{69.18} & \second{71.80} & \second{67.26} & \second{38.74} & \second{46.52} & \second{46.84} & \second{45.64} & 8.82 & 137.4
\\

& HOMER~\cite{pmlr-v205-patel23a}
& 46.72 & 67.54 & 69.84 & 64.46 & 37.31 & 44.59 & 44.98 & 43.62 & 9.29 & 145.9
\\

& \fm (ours)
& \best{57.70} & \best{71.15} & \best{72.62} & \best{69.26} & \best{45.31} & \best{49.52} & \best{49.72} & \best{48.95} & 9.01 & 141.1
\\
\midrule

% ------- Even -------
\multirow{9}{*}{\rotatebox{90}{\textcolor{hab2}{\textbf{Habit \#2}}}}
& Naive \ac{LLM}
& 39.02 & 58.52 & 60.66 & 55.46 & 31.30 & 37.95 & 38.21 & 36.94 & 8.72 & 135.2
\\

& TAP-LGX~\cite{dorbala2026personalizedembodiednavigationportable}
& 34.26 & 54.92 & 61.64 & 53.57 & 28.12 & 40.73 & 43.18 & 39.47 & 10.98 & 199.6
\\

& OSG+SP~\cite{10160473}
& 19.18 & 57.38 & 66.07 & 52.69 & 15.74 & 32.03 & 34.49 & 29.99 & 13.37 & 233.3
\\

& OSG+LLM~\cite{10160473}
& 36.89 & 47.38 & 50.16 & 46.43 & 29.85 & 34.97 & 35.69 & 34.40 & 8.14 & 129.5
\\

& CEG+SP~\cite{wang2024dynamic}
& 22.46 & 58.36 & 65.74 & 53.08 & 18.31 & 33.39 & 35.27 & 31.17 & 12.29 & 212.6
\\

& CEG+LLM~\cite{wang2024dynamic}
& 37.38 & 48.20 & 50.98 & 47.25 & 30.11 & 35.47 & 36.17 & 34.90 & 8.02 & 127.5
\\

& SGM~\cite{kurenkov2023modeling}
& \second{47.54} & 66.23 & \second{68.52} & \second{63.85} & \second{38.07} & \second{44.83} & \second{45.14} & \second{43.94} & \second{7.76} & \second{117.6}
\\

& HOMER~\cite{pmlr-v205-patel23a}
& 41.48 & \best{67.38} & \best{69.18} & 62.90 & 33.59 & 41.65 & 41.85 & 40.37 & 8.79 & 135.4 \\

& \fm (ours)
& \best{50.49} & \second{66.56} & 67.21 & \best{63.92} & \best{41.18} & \best{46.51} & \best{46.59} & \best{45.72} & \best{7.55} & \best{113.5}
\\
\midrule

%------- Instant -------

\multirow{9}{*}{\rotatebox{90}{\textcolor{hab3}{\textbf{Habit \#3}}}}
& Naive \ac{LLM}
& \second{41.15} & 56.23 & 61.48 & 55.46 & \second{33.52} & 38.31 & 39.21 & 37.89 & 9.45 & 145.1
\\

& TAP-LGX~\cite{dorbala2026personalizedembodiednavigationportable}
& 36.72 & 57.05 & 64.10 & 55.61 & 29.82 & \second{42.30} & \second{44.86} & \second{40.97} & 9.37 & 163.0
\\

& OSG+SP~\cite{10160473}
& 15.90 & 56.23 & 65.25 & 51.21 & 12.95 & 31.00 & 33.31 & 28.55 & 11.92 & 207.0
\\

& OSG+LLM~\cite{10160473}
& 38.52 & 54.26 & 55.74 & 52.33 & 31.72 & 38.97 & 39.33 & 38.04 & \best{7.68} & \best{115.7}
\\

& CEG+SP~\cite{wang2024dynamic}
& 18.52 & 57.21 & 64.43 & 52.21 & 14.85 & 31.48 & 33.30 & 29.37 & 11.54 & 199.2
\\

& CEG+LLM~\cite{wang2024dynamic}
& 38.03 & 54.10 & 55.74 & 52.00 & 31.23 & 38.59 & 38.93 & 37.62 & 7.91 & 119.9
\\

& SGM~\cite{kurenkov2023modeling}
& 27.21 & 62.62 & 65.41 & 56.84 & 22.19 & 33.64 & 34.02 & 31.84 & 12.24 & 198.1
\\

& HOMER~\cite{pmlr-v205-patel23a}
& 38.03 & \second{62.95} & \second{65.74} & \second{58.93} & 30.29 & 38.55 & 38.94 & 37.25 & 11.51 & 186.9
\\

& \fm (ours)
& \best{50.98} & \best{66.72} & \best{68.20} & \best{64.39} & \best{41.07} & \best{45.99} & \best{46.17} & \best{45.30} & \second{7.90} & \second{117.4}
\\
\bottomrule
\end{tabular}%
}

\caption{Results of \fm and the baselines for the dynamic \ac{ObjNav} episodes. We report metrics grouped by the three distinct habits that induce the objects displacement.}
\label{tab:objvav}
\end{table}

\begin{wrapfigure}[15]{r}{0.65\linewidth}
\centering
\includegraphics[width=\linewidth]{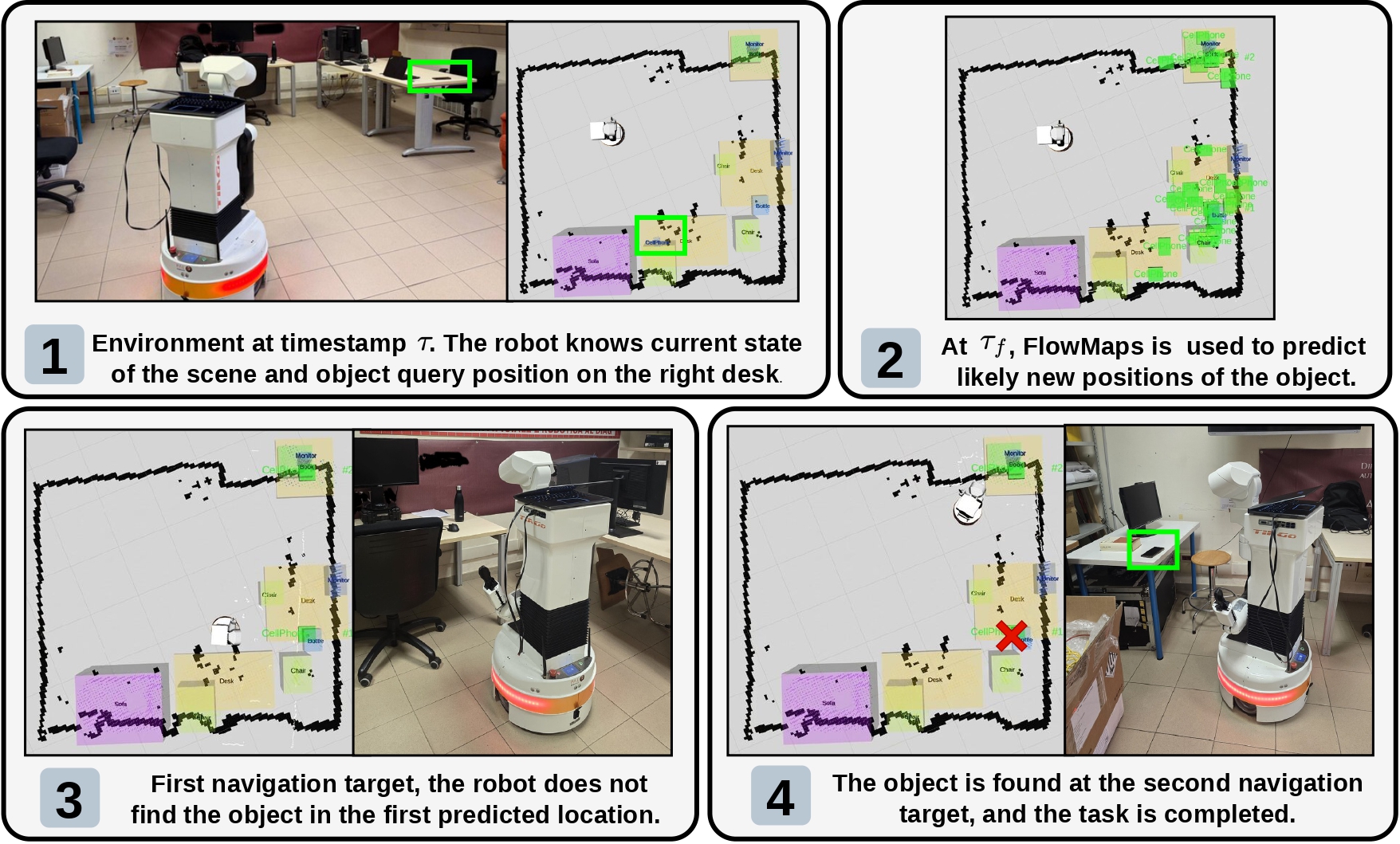}
\caption{\fm deployed in a real-world environment.}
\label{fig:exp}
\end{wrapfigure}

\boldparagraph{Real-world deployment} We further showcase our proposed approach in a real-world demonstration with a TIAGo robot\footnote{https://pal-robotics.com/robot/tiago/}. 
The experiment is depicted in Fig.~\ref{fig:exp}: it consists in finding a ``CellPhone" that belongs to a person operating in this environment and alternates between the different desks, spending an equal amount of time in each of them (\textcolor{hab2}{Habit~\#2}). 
The scene is observed when the phone is at the right desk. A couple of hours pass, and the robot is fetched to find again the phone. Predictions of \fm are clustered together and become ranked navigation targets for the agent. Eventually, the phone gets found at second targeted position.

\begin{comment}

DSG (CEG/OSG): classical score-based planning (prob × 1/dist)
LLM / TAP-LGX: text-based reasoning over history
HoMeR: GNN over a temporal graph of past placements
SGM: link prediction over a partially-observable scene graph

\end{comment}

\section{Limitations and Future Directions}
\label{sec:6_lfd}
The current formulation leaves room for further extensions. First, we consider a closed set of 41 object classes and 17 predefined receptacles, which makes \fm not directly applicable to open-vocabulary settings. This choice was motivated by our focus on indoor household environments, where relevant object classes and receptacles are relatively constrained. Extending the method to open-vocabulary labels could broaden its applicability to more diverse indoor scenarios, such as offices, laboratories, or factories. 
A second limitation concerns the amount of data needed to train the models. In our experiments, ProcTHOR enabled us to generate sufficient training data for the considered setup. However, extending the approach to more general environments would likely require comparable data, possibly without access to the same simulation tools. One direction to mitigate this issue is to reduce reliance on fully offline training. For example, after an appropriate sample-efficiency analysis, a compact offline backbone could be trained and then fine-tuned online using data collected during deployment.
Finally, the current evaluation is limited to scenarios with at most three habits. While sufficient to assess the proposed formulation, further analysis is needed to understand how the method scales with the number and diversity of habits.
These limitations will be investigated and addressed in future work.
\section{Conclusions}
\label{sec:7_conc}
In this paper, we introduced \fm, a latent \ac{FM} framework for modeling multimodal spatio-temporal distributions of dynamic objects in human-inhabited environments. Rather than predicting discrete receptacle-level states, \fm models object relocalization as a distribution over future bounding boxes in continuous 3D space. Conditioned on the current scene, queried object, and prediction horizon, it learns recurring human-driven motion patterns directly from data, without explicit habit or activity labels. By combining a \ac{VAE}-based latent object representation with a \ac{CDiT} flow model, \fm remains grounded in object geometry while generalizing to unseen household layouts. We evaluated both its distributional properties and practical utility in dynamic \ac{ObjNav}, where predicted future-location distributions generate ranked navigation targets. Across ProcTHOR simulations and real-world robotic deployments, \fm captures continuous, multimodal, and cross-environment spatio-temporal structure, improving object search and providing an effective prior for robotic reasoning beyond static scene assumptions.
\acknowledgments{This work has been carried out while Francesco Argenziano was enrolled in the Italian National Doctorate on Artificial Intelligence run by Sapienza University of Rome. This research was conducted while Francesco Argenziano was enrolled as a visiting researcher at Mila - Quebec AI Institute. The work conducted at the Université de Montréal was supported by the Natural Sciences and Engineering Research Council of Canada (NSERC) through an NSERC Discovery Grant and by the CIFAR AI Chair program (Liam Paull). Individual support was provided through NSERC Postgraduate Scholarships-Doctoral (PGS D) (Sacha Morin, Charlie Gauthier, Miguel Saavedra-Ruiz). This research was also enabled in part by computational resources provided by Mila (mila.quebec).}
\clearpage
\appendix
\newcolumntype{Y}{>{\raggedright\arraybackslash}X}
{\LARGE \textbf{Appendix}}
\label{app:dataset}
\begin{table}[h!]
\centering
\caption{Closed-set semantic vocabularies used for dataset generation.}
\label{tab:appendix-class-vocabularies}
\scriptsize
\setlength{\tabcolsep}{3pt}
\renewcommand{\arraystretch}{0.95}
\begin{tabular*}{\linewidth}{@{\extracolsep{\fill}} rlrlrlrlrl @{}}
\toprule
\multicolumn{6}{c}{\textbf{Pickupable objects} $\mathcal{O}_{\mathrm{obj}}$}
&
\multicolumn{4}{c}{\textbf{Receptacles} $\mathcal{O}_{\mathrm{BG}}$}
\\
\cmidrule(lr){1-6}
\cmidrule(lr){7-10}
ID & Class & ID & Class & ID & Class & ID & Class & ID & Class \\
\midrule
0  & Apple        & 14 & Bottle        & 28 & Spoon          & 0  & CounterTop   & 9  & Sink \\
1  & Egg          & 15 & Knife         & 29 & Watch          & 1  & Plate        & 10 & Sofa \\
2  & Fork         & 16 & Spatula       & 30 & BaseballBat    & 2  & Pot          & 11 & TVStand \\
3  & Kettle       & 17 & Bread         & 31 & BasketBall     & 3  & ShelvingUnit & 12 & GarbageCan \\
4  & Ladle        & 18 & ButterKnife   & 32 & DishSponge     & 4  & Chair        & 13 & SideTable \\
5  & Lettuce      & 19 & Candle        & 33 & ToiletPaper    & 5  & DiningTable  & 14 & Desk \\
6  & Pencil       & 20 & CellPhone     & 34 & TissueBox      & 6  & Dresser      & 15 & CoffeeTable \\
7  & Potato       & 21 & Newspaper     & 35 & TeddyBear      & 7  & ArmChair     & 16 & Box \\
8  & SaltShaker   & 22 & Vase          & 36 & SoapBar        & 8  & Bed          &    & \\
9  & SoapBottle   & 23 & AlarmClock    & 37 & PaperTowelRoll &    &              &    & \\
10 & SprayBottle  & 24 & KeyChain      & 38 & DeskLamp       &    &              &    & \\
11 & Tomato       & 25 & Laptop        & 39 & TennisRacket   &    &              &    & \\
12 & WineBottle   & 26 & Pen           & 40 & Cloth          &    &              &    & \\
13 & Book         & 27 & RemoteControl &    &                &    &              &    & \\
\bottomrule
\end{tabular*}
\end{table}
\section{Dataset and environments}
\boldparagraph{Environment classes}
We generated the environments using ProcTHOR and adopted the closed set of semantic labels available in the simulator for dynamic pickupable objects and receptacles. This resulted in a vocabulary $\mathcal{O}_{\mathrm{obj}}$ of $41$ pickupable object classes and a vocabulary of $\mathcal{O}_{\mathrm{BG}}$ $17$ receptacle classes. The former contains the dynamic objects whose locations may change over time, while the latter contains the static receptacles and furniture elements that define the nvironment layout and provide support surfaces for object placement. The complete list can be observed in Table~\ref{tab:appendix-class-vocabularies}.

\begin{figure}[h!]
    \centering
    \includegraphics[width=\linewidth]{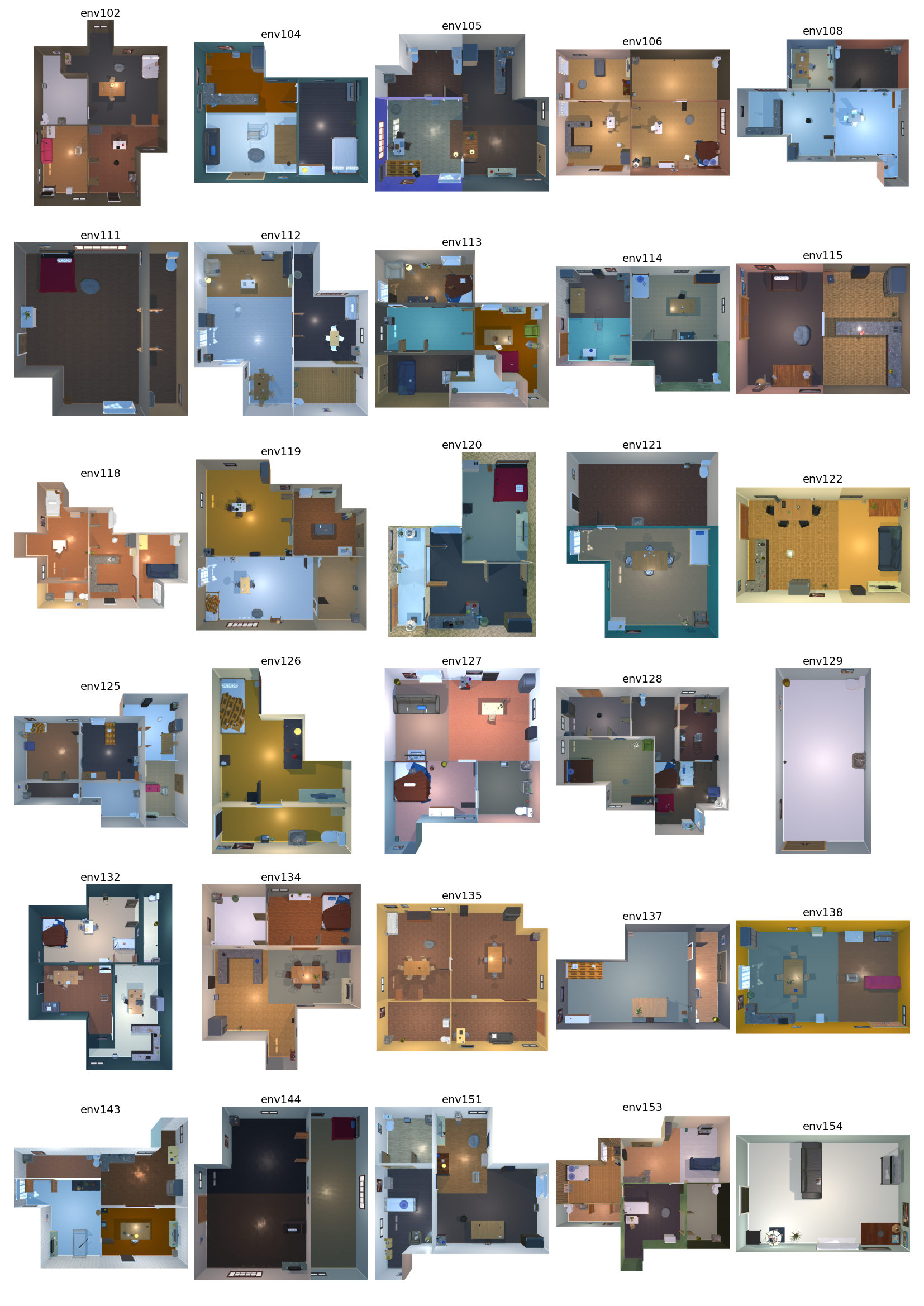}
    \caption{An overview of 30 ProcTHOR environments coming from the validation split.}
    \label{app:topdown}
\end{figure}

\boldparagraph{Scene layouts} The generated dataset contains indoor ProcTHOR scenes with diverse spatial layouts and room configurations. Scenes range from compact single-room environments to larger multi-room
households, with different floor plans, furniture arrangements, and numbers of receptacle instances. This variability affects both the geometric structure of the environment and the set of valid support surfaces available for object placement. A sample of 30 validation environments can be observed in Fig.~\ref{app:topdown}

\begin{wrapfigure}[16]{r}{0.7\linewidth}
    \centering
    \vspace{-1.0em}
    \includegraphics[width=\linewidth]{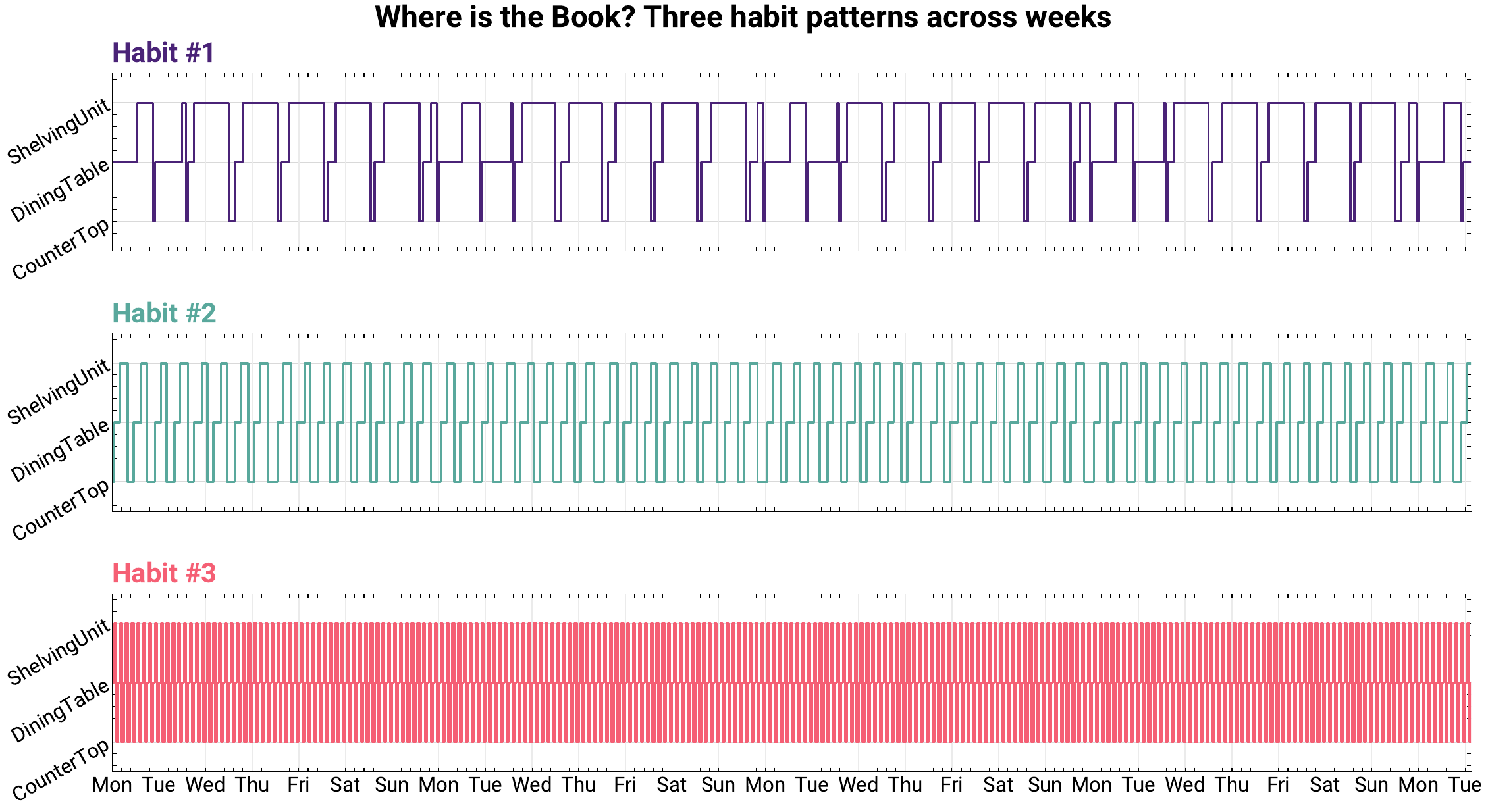}
    \caption{Different spatio-temporal behaviors of the same object (``Book") under different human habits in a ProcTHOR scene.}
    \label{fig:habits}
\end{wrapfigure}
\boldparagraph{Object-receptacle compatibility and habit-conditioned schedules}
To avoid arbitrary or semantically invalid placements, we use ProcTHOR's pickupable and receptacle weighting files to determine which pickupable object classes can be assigned to which receptacle classes.
These metadata define a class-level compatibility prior between objects and support surfaces, ensuring that each object is placed only on semantically plausible receptacles. For example, kitchen objects such as Fork or Tomato may be assigned to counters, sinks, plates, or dining tables, while unrelated receptacles are excluded from their candidate sets. For each object class $o \in \mathcal{O}_{\mathrm{obj}}$, we define the compatible receptacle set as $\mathcal{R}(o) \subseteq \mathcal{O}_{\mathrm{BG}}$. Then, for each object instance in a scene, we select candidate receptacle instances whose class belongs to $\mathcal{R}(o)$. \newpage
Each dynamic scene is generated by assigning every pickupable object a habit-conditioned trajectory over this candidate set. The compatibility 
prior constrains which receptacles are valid, while the habit determines the temporal structure of the object's receptacle sequence. We consider the three habit families described in the main text. \textcolor{hab1}{Habit\#1} captures a preference-driven routine, where the simulated human spends more time in a few favored locations. \textcolor{hab2}{Habit\#2} represents a balanced routine, with time distributed roughly evenly across relevant locations. \textcolor{hab3}{Habit\#3} models a more dynamic routine, characterized by frequent transitions and short stays. An example of such rouines can be observed in Fig.~\ref{fig:habits}.

\boldparagraph{Continuous placement by potential-field sampling} The habit schedule specifies the target receptacle for each object at each timestamp, but not a single deterministic 3D placement. To obtain continuous object locations, we sample placements from a potential field defined over the support surface of the selected receptacle. The sampled surface point determines the continuous position of the object on the receptacle, and the object is then teleported to the corresponding pose in the simulator. This separates the discrete semantic component of the data generation process from the continuous geometric one: the habit determines which receptacle an object moves to, while the potential field determines where on that receptacle surface the object is placed. This procedure produces trajectories in which objects follow semantically consistent, habit-conditioned patterns while still exhibiting continuous spatial variation. As a result, the model is not trained to predict only discrete object-receptacle relations, but rather full future bounding boxes in continuous 3D space.

\section{Implementation Details}
\label{app:impl}

\subsection{Tokenization and encoding strategies}
\label{app:tokens}
\boldparagraph{Bounding box normalization}
Because ProcTHOR generates scenes with varying layouts, we normalize all bounding box center coordinates to $[0,1]$ using fixed global bounds, $c_\text{min}$ and $c_\text{max}$, computed from the extrema observed in the training set. Since object sizes vary substantially across semantic categories, for example between an ``Egg'' and a ``Bed'', we map bounding box extents to log-space and normalize them using the observed log-size range in the training data. Scenes are then tokenized to a fixed sequence length $S = 32$, as no environment exceeded this number of receptacles plus dynamic objects.

\boldparagraph{Embeddings} Each normalized bounding box is embedded through four separate branches, corresponding to translation magnitude, translation direction, size magnitude, and size direction. Each branch is a 4-layer \ac{MLP}, implemented as two stacked $\text{Linear}\!\to\!\text{GELU}\!\to\!\text{Linear}$ blocks with hidden width $d = 256$. Translation- and size-group sums are individually layer-normalized, summed, and passed through a final layer norm to produce the per-token geometric embedding. Semantic labels and the dynamic/background flag are embedded using learned tables of width $d$, each followed by a 2-layer $\text{Linear}\!\to\!\text{SiLU}\!\to\!\text{Linear}$ \ac{MLP}. Receptacle and dynamic-object classes use \emph{separate} embedding tables to reflect their different cardinalities and priors. Hyperparameters are summarized in Table~\ref{tab:app-tokens}.

\begin{table}[t]
  \centering
  \small
  \caption{Tokenization and bounding box encoder hyperparameters.}
  \label{tab:app-tokens}
  \begin{tabularx}{\linewidth}{@{}p{0.50\linewidth}Y@{}}
    \toprule
    \textbf{Parameter} & \textbf{Value} \\
    \midrule
    Sequence length $S$                                           & $32$ \\
    Hidden width $d$                                              & $256$ \\
    Center normalization range $[c_{\min}, c_{\max}]$             & $[-0.5, 34.0]\,\mathrm{m}$ \\
    Log-size offset $\varepsilon_s$                               & $10^{-4}$ \\
    Bounding box branches                                         & $4$ translation/size $\times$ direction/magnitude branches \\
    Bounding box MLP depth                                        & $4$ layers, GELU \\
    Label/flag MLP depth                                          & $2$ layers, SiLU \\
    Linear init, bounding box branches and label MLPs              & $\mathcal{N}(0, 0.02^2)$, bias $0$ \\
    LayerNorm init                                                & gain $1$, bias $0$ \\
    \bottomrule
  \end{tabularx}
\end{table}

\subsection{Variational Autoencoder}
\label{app:vae}

\begin{wrapfigure}[14]{r}{0.7\linewidth}
    \centering
    \vspace{-0.7em}
    \includegraphics[width=\linewidth]{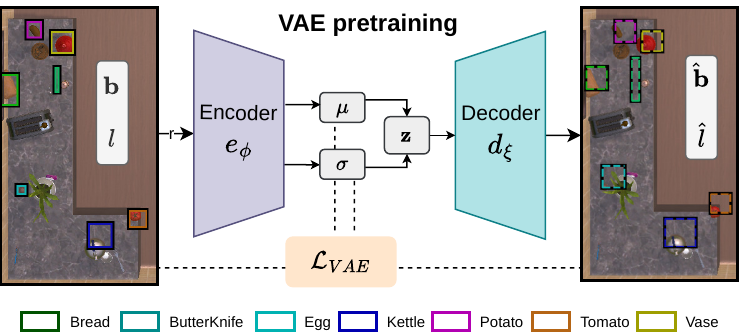}
    \caption{\ac{VAE} used to encode and decode object tokens in a ProcTHOR environment. Dotted bounding boxes mean decoded objects.}
    \label{fig:vae}
    %\vspace{-1em}
\end{wrapfigure}

A general overview of the \ac{VAE} can be observed in Fig.~\ref{fig:vae}. The encoder $e_\phi$ and decoder $d_\xi$ are residual \acp{MLP} with hidden
width $d_{\text{vae}} = 256$, depth $L_{\text{vae}} = 3$, and an \ac{MLP}
expansion ratio of $4$. Each residual block is a
$\text{Linear}\!\to\!\text{LayerNorm}\!\to\!\text{SiLU}\!\to\!
\text{Dropout}\!\to\!\text{Linear}\!\to\!\text{LayerNorm}\!\to\!
\text{SiLU}\!\to\!\text{Dropout}$ stack with a skip connection. The
encoder input is the bounding box embedding summed with the class-label
embedding described in Section~\ref{app:tokens}; its final linear layer
projects to $2d_z$ and is split into
$(\boldsymbol{\mu},\,\log\boldsymbol{\sigma}^2)$. The decoder mirrors
the encoder and terminates in two heads: a 6-dimensional geometry head
with sigmoid activation, so outputs stay in $[0,1]^6$, and a
$|\mathcal{O}_{\text{obj}}|$-way class-logit head.
The CIoU term is computed in metric space after inverting the log-size mapping, so that the area/volume penalty reflects actual overlap rather than overlap in the normalised representation. After training, per-dimension latent statistics
$(\boldsymbol{\mu}_{\mathcal{D}}, \boldsymbol{\sigma}_{\mathcal{D}})$ are computed once on the training split and cached on disk. The \ac{FM} stage operates entirely on standardized latents $(\mathbf{z} - \boldsymbol{\mu}_{\mathcal{D}}) / \boldsymbol{\sigma}_{\mathcal{D}}$, with the inverse transform applied to predicted latents before decoding. All other architectural and optimization hyperparameters are reported in Table~\ref{tab:app-vae}.

\begin{table}[t]
  \centering
  \small
  \caption{VAE hyperparameters.}
  \label{tab:app-vae}
  \begin{tabularx}{\linewidth}{@{}p{0.50\linewidth}Y@{}}
    \toprule
    \textbf{Parameter} & \textbf{Value} \\
    \midrule
    Latent dimension $d_z$                                      & $32$ \\
    Hidden width $d_{\text{vae}}$                               & $256$ \\
    Depth $L_{\text{vae}}$                                      & $3$ \\
    MLP expansion ratio                                         & $4$ \\
    Dropout                                                     & $0.1$ \\
    Linear init                                                 & Xavier-uniform, bias $0$ \\
    \midrule
    Iterations                                                  & $30{,}000$ \\
    Batch size                                                  & $256$ \\
    Optimizer                                                   & AdamW \\
    Learning rate                                               & $10^{-4}$ \\
    $(\beta_1, \beta_2)$                                        & $(0.95, 0.999)$ \\
    $\varepsilon$                                               & $10^{-8}$ \\
    Weight decay                                                & $10^{-6}$ \\
    LR schedule                                                 & cosine, $500$ linear-warmup steps \\
    \midrule
    $\lambda_{\text{CIoU}}$                                     & $1.0$ \\
    $\lambda_{L_1}$                                & $5.0$ \\
    $\lambda_{\text{CE}}$                                       & $1.0$ \\
    Final KL weight $\beta$                                     & $10^{-2}$ \\
    KL warm-up steps $T_{\text{kl}}$                            & $10{,}000$, linear from $0$ to $\beta$ \\
    \bottomrule
  \end{tabularx}
\end{table}

\subsection{Latent Flow Matching Network}
\label{app:fm}

\boldparagraph{Map encoder}
The map-encoder layers are pre-norm Transformer Encoder modules with GELU activations. The dynamic/background flag is implemented as a 2-entry learned parameter, with one $d$-dimensional vector per type, added on top of the bounding box, class, and observation time embeddings before the first layer.

\boldparagraph{CDiT block}
The adaLN-Zero projection is a single SiLU \(\rightarrow\) Linear\((d,8d)\) MLP whose output
is split into eight chunks \((\gamma_1,\beta_1,\alpha_1,\gamma_2,\beta_2,\alpha_2,\gamma_3,\beta_3)\).
The first triple modulates the query token and gates the cross-attention residual.
The second triple modulates and gates the feed-forward sublayer. The last pair modulates the context tokens, namely keys and values, before the cross-attention
sublayer. The feed-forward sublayer is
a two-layer Linear \(\rightarrow\) GELU \(\rightarrow\) Linear MLP with the same expansion ratio 4
as the map encoder. Cross-attention uses \(H\) heads with biases on the key/value projections.

\boldparagraph{Conditioning embeddings}
The flow time $t \in [0,1]$ is encoded with a $256$-dimensional
sinusoidal frequency embedding followed by a 2-layer
$\text{Linear}\!\to\!\text{SiLU}\!\to\!\text{Linear}$ MLP. The final
timestamp $\tau_f$ and the observation time $\tau$ use \emph{separate}
learned tables of size $T_{\max} \times d$. The queried object label
$l_q$ uses its own 2-layer-MLP class embedder, distinct from the
map-encoder object table, so that the query and map tokens are not tied.
The conditioning vector fed to each CDiT block is the sum of the
embeddings of $t$, $\tau_f$, and $l_q$. The noisy latent $\mathbf{z}_t$
is lifted from $d_z = 32$ to $d = 256$ by a 2-layer
$\text{Linear}\!\to\!\text{SiLU}\!\to\!\text{Linear}$ input embedder.

\boldparagraph{Initialization}
All adaLN-Zero projections and the gate projection of the final output head are zero-initialized so that the network starts as an identity map on the noisy latent~\cite{peebles2023scalable}. Other linear layers are initialized with Xavier-uniform. Embedding tables and the projection MLPs of the class and timestep embedders use $\mathcal{N}(0, 0.02^2)$. Architectural hyperparameters are reported in Table~\ref{tab:app-fm}.

\begin{table}[t]
  \centering
  \small
  \caption{Latent flow-matching network hyperparameters.}
  \label{tab:app-fm}
  \begin{tabularx}{\linewidth}{@{}p{0.50\linewidth}Y@{}}
    \toprule
    \textbf{Parameter} & \textbf{Value} \\
    \midrule
    Hidden width $d$                                        & $256$ \\
    Attention heads $H$                                     & $8$ \\
    MLP expansion ratio                                     & $4$ \\
    Map encoder depth $N_T$                                 & $8$ \\
    CDiT block depth $N_C$                                  & $8$ \\
    Dropout                                                 & $0.1$ \\
    \midrule
    Latent input embedder                                   & 2-layer MLP, $d_z \!\to\! d$, SiLU \\
    Flow time embedder                                      & sinusoidal, $256$, plus 2-layer MLP \\
    $\tau$, $\tau_f$ embedders                              & separate $T_{\max} \times d$ tables \\
    Query label embedder                                    & 2-layer MLP, separate table \\
    Conditioning vector $c$                                 & $\mathrm{emb}(t) + \mathrm{emb}(\tau_f) + \mathrm{emb}(l_q)$ \\
    \midrule
    Init: adaLN-Zero and final gate                         & zero \\
    Init: other linear layers                               & Xavier-uniform \\
    Init: embedding tables, class/time MLPs                 & $\mathcal{N}(0, 0.02^2)$ \\
    \bottomrule
  \end{tabularx}
\end{table}

\subsection{FlowMaps Training and Sampling}
\label{app:hp}
We adopt the CondOT path~\cite{lipmanflow} with $\sigma_{t}=0$, which reduces the conditional probability path to the straight-line interpolation $\mathbf{z}_t = (1-t)\mathbf{z}_0 + t\mathbf{z}_1$ with constant target velocity $u_t^{\text{target}} = \mathbf{z}_1 - \mathbf{z}_0$. Within each mini-batch, source and target samples are paired by an exact Hungarian solver on the squared-Euclidean cost $\|\mathbf{z}_0 - \mathbf{z}_1\|_2^2$. Only $\mathbf{z}_0$ is permuted, leaving $\mathbf{z}_1$ and every conditioning variable in their original order so that semantic alignment with $(M_\tau, l_q, \tau_f)$ is preserved. We sample $t = \mathrm{sigmoid}(s)$ with $s \sim \mathcal{N}(0, 1)$, giving a logit-normal flow-time distribution. This concentrates supervision on the middle of the trajectory, where the target velocity has its largest variance. We also experimented with sampling $t$ from a uniform distribution $U[0,1]$ and from a Beta distribution, but empirically found the logit-normal schedule to yield the best results. During inference, we integrate with a fixed-step Euler solver on a uniform grid over $[0,1]$, with a step of $0.05$. Smaller steps and different intregration methods (\eg midpoint) produced no measurable improvement. The map encoder is run once per scene and its output reused across all dynamic objects in the scene and across the $N_{\text{pred}}$ posterior samples drawn for each query. Training and sampling hyperparameters are summarized in Table~\ref{tab:app-hp}.

\begin{table}[t]
  \centering
  \small
  \caption{Flow-matching training and sampling hyperparameters.}
  \label{tab:app-hp}
  \begin{tabularx}{\linewidth}{@{}p{0.50\linewidth}Y@{}}
    \toprule
    \textbf{Parameter} & \textbf{Value} \\
    \midrule
    Iterations                                      & $30{,}000$ \\
    Batch size                                      & $1024$ \\
    Optimizer                                       & AdamW \\
    Learning rate                                   & $10^{-4}$ \\
    $(\beta_1, \beta_2)$                            & $(0.95, 0.999)$ \\
    Weight decay                                    & $10^{-6}$ \\
    LR schedule                                     & cosine, $500$ linear-warmup steps \\
    Validation interval                             & every $2{,}500$ iterations \\
    \midrule
    Path                                            & CondOT~\cite{lipmanflow}, $\sigma = 0$ \\
    Coupling                                        & exact mini-batch OT, Hungarian \\
    Flow-time sampler                               & logit-normal, $s \sim \mathcal{N}(0, 1)$ \\
    Loss                                            & squared-$L_2$ CFM regression \\
    \midrule
    EMA decay                                       & $0.9999$ \\
    EMA inverse-gamma / power                       & $1.0$ / $0.75$ \\
    EMA warm-up                                     & none \\
    \midrule
    ODE solver, inference                           & euler  \\
    Step size                                       & $0.05$ \\
    \bottomrule
  \end{tabularx}
\end{table}

\section{Experiments}
\subsection{Distributional Evaluation: Experimental Details}
\label{app:distributional}

We evaluate on a shared minival split of 25 validation environments, identical across the three habits described in the main paper. For each scene with $T$ hourly timestamps we select $n_{\tau}$ evenly-spaced observation times $\tau \in [0, T{-}2]$ and, for each $\tau$, up to $10$ evenly-spaced query times $\tau_f \in (\tau, T{-}1]$, giving at most $30$ $(\tau, \tau_f)$ pairs per (env, $l_q$). For each pair we draw $K{=}50$ samples from the predictor. Per-pair, per-instance \emph{minFDE@K} and \emph{Rec@1} are then averaged over all $(\tau, \tau_f)$ pairs and all instances of class $l_q$ in the environment. In contrast, the distributional shape metrics are computed once per $(\mathrm{env}, l_q)$ using pooled point clouds. The predicted cloud is obtained by concatenating the $K$ sampled centers across all pairs, yielding $M \approx 1500$ points. The ground-truth cloud is the union of the bounding box centers of every instance of class $l_q$ at every evaluated $\tau_f$.

Regarding the metrics, \emph{minFDE} uses the full 3D Euclidean distance to the ground-truth bounding box center, while \emph{Rec@1} uses xz-only distance with $\delta{=}1m$. \emph{Density} and \emph{Coverage} follow \citet{pmlr-v119-naeem20a} with $k_{\mathrm{NN}}{=}5$, augmented with a $0.5m$ radius floor that prevents degenerate $k$-NN radii on classes whose ground-truth positions coincide across timestamps (\eg static objects on the same receptacle). \emph{TV} and \emph{JS} are computed on an xz floor grid of resolution $0.5$\,m, with JS reported in nats (bounded by $\ln 2 \approx 0.693$). We rely on sample-based histogram divergences rather than the change-of-variables likelihood obtainable from the flow itself for two distinct reasons. First, the inverse-ODE likelihood is defined only for \fm, and even there only in the \ac{VAE} latent space the flow operates on; the \ac{VAE} decoder is not invertible, so a likelihood in 3D position units would require an importance-weighted estimate that introduces its own variance and bias. The baselines fare no better: FreqPrior and LLMPrior have closed-form densities, but those are two-dimensional delta measures concentrated on receptacle top surfaces, while EmpiricalMean and Mean\fm are point masses. In every, case the pointwise log-likelihood at a generic 3D ground-truth coordinate is either $-\infty$ or ill-defined, and the only way to make the numbers comparable is to fit a separate density estimator (\eg kernel density estimation) on each method's samples, which introduces additional hyperparameters and confounds the comparison. Second, even within \fm the inverse-ODE likelihood scores how much density the model assigns at the ground-truth coordinates, whereas the downstream use of \fm consumes \emph{samples} drawn from the model rather than its evaluated density: a metric defined on the predicted samples is therefore more directly aligned with the actual object of comparison. TV and JS satisfy both requirements simultaneously: they are computable from any baseline's samples without further fitting, and they measure how the predicted mass is \emph{spatially allocated} across the scene rather than how dense it is at the reference point.

\boldparagraph{FreqPrior} This baseline represents a class-conditional categorical over receptacle categories. From the training split of the current habit we estimate, for each pickupable class $l_q$ and receptacle category $r$, the empirical frequency of $l_q$ resting on instances of $r$, and store the resulting table $P(r \mid l_q)$ alongside the data. At test time, for each of the $K$ samples we \emph{(i)} restrict $P(\cdot \mid l_q)$ to receptacle categories present in $M_\tau$ and renormalize, \emph{(ii)} sample a category $r$, emph{(iii)} sample a concrete receptacle of that category uniformly, and \emph{(iv)} sample a point uniformly inside its footprint polygon. The prior is time-invariant and the scene receptacles are read at $\tau$ and reused for any $\tau_f$, so this baseline has no notion of time.

\boldparagraph{LLMPrior} LLMPrior uses the same point-sampling pipeline as FreqPrior, but the distribution over receptacle \emph{instances} is replaced by a ranked top-$T$ list suggested from an \ac{LLM}. We use \texttt{llama3.1:8b} as our \ac{LLM}, with $T{=}5$, sampling temperature $0$, and a rank-softmax temperature of $1.0$, so the rank-$i$ receptacle receives mass proportional to $\exp(-i)$. One LLM call is issued per $(\text{env}, l_q, \tau, \tau_f)$: the returned indeces induce the receptacle distribution, and the $K$ predicted points are obtained by drawing receptacles from it and sampling uniformly inside the chosen footprint at its top face. The system prompt remains unchanged across queries. For each query, we generate the user prompt by filling the bold-bracketed placeholders in the box below. These placeholders specify the observation time $\tau$, the scene state $M_\tau$, the queried class $l_q$, the receptacle that currently contains the target, the elapsed time $\tau_f - \tau$, the query time $\tau_f$, and the top-$T$ size. The scene state is written as a bullet list over receptacles, with each line reporting the pickupable categories currently on that receptacle, or ``empty'' when none are present. Timestamps are formatted as \texttt{Week W, Weekday HH:00} starting from \texttt{Week 1, Monday 00:00}, and elapsed time as \texttt{X day(s) and Y hour(s)}.

\begin{tcolorbox}[
  colback=orange!4!white, colframe=black!55!white,
  sharp corners, boxrule=0.4pt,
  left=5pt, right=5pt, top=5pt, bottom=5pt,
]
\small
\textcolor{red!65!black}{\textbf{System Prompt:}}
You are an agent that has to find an object that may have moved from its previous location. You will be given the layout of receptacles in a scene and the object's previous location. Reason about where the object is most likely to be now.\\[4pt]
\textcolor{blue!65!black}{\textbf{User Prompt:}}
You are an agent that has to find an object that could have moved from its previous location. You know that on \textbf{[$\tau$]} the environment has the following layout: \textbf{[$M_\tau$ receptacle list]}. Also, on \textbf{[$\tau$]}, the \textbf{[$l_q$]} is on \textbf{[current receptacle of $l_q$]}. \textbf{[$\tau_f - \tau$]} later, it is now \textbf{[$\tau_f$]}. From the list of receptacles above, enumerate the top \textbf{[$T$]} most likely receptacles where the \textbf{[$l_q$]} can be found on \textbf{[$\tau_f$]}, ordered from most to least likely. Reply with one receptacle id per line, exactly as written in the list above. No commentary, no numbering, no extra text.
\end{tcolorbox}

\noindent
When the queried object is not assigned to any receptacle at $\tau$, the ``\textbf{[$l_q$]} is on \textbf{[current receptacle of $l_q$]}'' clause is replaced by ``the \textbf{[$l_q$]} is not on any of the listed receptacles''.

\boldparagraph{EmpiricalMean} This baseline represents an oracle point predictor: for each $(\text{env}, l_q)$, it returns the centroid of every ground-truth bounding-box center of class $l_q$ pooled across the scan's timestamps, replicated $K$ times. The predictor consults the evaluation set and is independent of $\tau$ and $\tau_f$; since this centroid minimizes $L_2$ error against the pooled ground-truth cloud, it upper-bounds any deterministic regression baseline conditioned on at least $(M_\tau, l_q)$. Its distribution-shape entries in the main results table are not reported since these metrics are not informative for a predictor that emits $K$ identical samples.

\boldparagraph{Mean\fm} This baseline is a wrapper around the trained \fm predictor that, per query, draws $K_{\mathrm{int}}{=}50$ samples internally, returns their arithmetic mean, and replicates that point $K$ times in the output. The conditioning $(M_\tau, l_q, \tau_f)$ is inherited from \fm; the only operation removed is multimodality, which isolates its contribution to the metric gains. Its distribution-shape entries are not reported for the same reason as EmpiricalMean.

\subsection{Object navigation: experimental details}
\label{app:objnav}

We evaluate on the same minival split of 25 validation environments used in the distributional protocol, with the agent embodied in AI2-THOR through ProcTHOR scene assets. For every environment we sample 5 target objects, and for each target we draw 5 $(\tau, \tau_f)$ pairs (the observation timestamp at which the agent perceives the scene, and the future timestamp at which we ask it to retrieve the object), yielding $600$+ episodes per habit. Each method receives, at the start of an episode, the per-timestep data that backs the scene up to $\tau$ (the observation history of receptacle-object assignments) and is asked to produce a ranked list of $K$ candidate future receptacles for the target at $\tau_f$. The agent then navigates the AI2-THOR controller toward those candidates in order, querying \texttt{GetShortestPathToPoint} for the planner-relevant distances, with a per-episode budget of $1000$ environment steps. An episode is successful at depth K if any of the first K candidates places the agent within $\delta_{\text{dist}}=1.25m$s of the target and the target is observed from the agent's point of view. In simulator runs, observation is assessed using AI2-THOR ground-truth visibility. In the real-robot demonstration, the VLM is used only to trigger a candidate target observation; a trial is counted as successful only when this trigger occurs at a valid target location, so VLM false positives are not counted as successful detections.

For \fm, the ranked candidate list is built by drawing $N_{\mathrm{preds}}{=}25$ samples per query from the trained flow, clustering them with DBSCAN ($\varepsilon{=}1.0$\,m, $\min{=}2$) and ranking clusters by mass. Ties are broken by the cluster centroid that minimises the distance from the current agent pose. Across all baselines we report Success Rate, SPL, and their per-$K$ variants exactly as in the main results table.

\boldparagraph{TAP-LGX}
We follow the formulation of \cite{dorbala2026personalizedembodiednavigationportable}, which extends the zero-shot exploration of LGX~\cite{dorbala2023can} with a rolling \emph{transit memory} of past observations. At each LGX step the agent performs a $360^\circ$ scan, the visible objects are fed into a VLM together with the current memory, the VLM returns a single object name, and the agent advances a fixed distance toward it; failures and target sightings are appended to the memory. We reproduce the published exploration loop in full: the LGX step-distance is $2m$, the per-episode budget is $30$ LGX steps, and the memory horizon is the last $30$ entries; on lock-on the agent navigates directly to the detected target. We deviate from the original setup in three points. First, we substitute the VLM with the same \texttt{llama3.1:8b} used elsewhere in our evaluation, constraining its output to the visible-object set with a JSON schema so the language model behaves as a discrete chooser instead of a free-form generator.Second, we replace the RGB-frame YOLO detections used in the original paper with ground-truth visibility queries from AI2-THOR, filtered to the LGX vocabulary. This is a substantial strengthening of the perception input, since it removes detector noise and gives the system the same perception-noise floor used by the other baselines in the table. Third, we provide input-parity to the other basilines, we pre-seed the rolling memory with the full per-receptacle scene state at every past timestep $t \le \tau$, and grows during exploration with one entry per LGX step recording the agent's pose, the list of visible objects of the current scan, and the heading the \ac{LLM} chose. The system and user prompts are shown in the box below: the system prompt renders the rolling memory verbatim, the user prompt is issued once per LGX step to elicit a single visible-object choice, and the bold-bracketed placeholders are filled per query with the target class $l_q$, the current memory contents, and the set of objects detected at the current scan.

\begin{tcolorbox}[
  colback=orange!4!white, colframe=black!55!white,
  sharp corners, boxrule=0.4pt,
  left=5pt, right=5pt, top=5pt, bottom=5pt,
]
\small
\textcolor{red!65!black}{\textbf{System Prompt:}}
I am a smart robot trying to find a \textbf{[$l_q$]} in my house. The target sometimes moves between receptacles over time; the lines below record where I have seen it or the scene before, and what I observed as I walked around. Use this to reason about where it likely is now.\\
Memory:\\
\textbf{[rolling memory: one line per past entry, e.g.\ ``[step $s$] near (x, z) heading $h$ deg observed [obj list] $\rightarrow$ went toward $a$'' or ``[prior $t$] target on $r$''; pre-seeded scene-state lines in TAP-LGX]}\\[4pt]
\textcolor{blue!65!black}{\textbf{User Prompt:}}
I want to find a \textbf{[$l_q$]} in my house. Which object from \textbf{[visible objects at the current scan]} should I go towards? Reply in ONE word.
\end{tcolorbox}

\boldparagraph{CEG}
The Cost-Effective Greedy planner of \citet{wang2024dynamic} can be viewed as a multiplicative simplification of the original OSG additive blend described below. It ranks unvisited receptacles by the ratio $p_i / d_i$, where $p_i$ is the estimated probability that receptacle $i$ currently contains the target, and $d_i$ is the navigation distance from the agent to that receptacle. At each step, the agent visits the highest-ranked receptacle. If the target is not found, the corresponding probability mass is set to zero, the remaining masses are renormalized, and the procedure continues until the target is found, all receptacles have been visited, or the per-episode receptacle budget is exhausted. We set this budget to $10$, matching the typical scene size. Navigation distances are computed using AI2-THOR's \texttt{GetShortestPathToPoint}. To avoid numerical artifacts when receptacles are collocated or nearly collocated, we use a $0.25m$ epsilon in the denominator of $p_i / d_i$, matching the ProcTHOR grid resolution. We evaluate two variants of the probability estimate $p_i$. The \emph{scene-prior} (SP) variant uses the per-class receptacle co-occurrence table $P(r \mid l_q)$, which is the same table used by FreqPrior in the distributional evaluation. The \emph{LLM} variant instead queries \texttt{llama3.1:8b} for an unnormalized probability for each receptacle. The \ac{LLM} estimator prompt, shown in the box below, follows the structure of the paper, but replaces the ``simulated resident'' human-activity hint with one of three habit-specific lines.
\begin{itemize}
\raggedright

\item \textcolor{hab1}{Habit~\#1}: \texttt{``This [\textbf{$l_q$}] has a few favorite places. Most often on [\textbf{$r_1$}] ([\textbf{$p_1$}]\%), sometimes [\textbf{$r_2$}] ([\textbf{$p_2$}]\%), and occasionally [\textbf{$r_3$}] ([\textbf{$p_3$}]\%).''}

\item \textcolor{hab2}{Habit~\#2}: \texttt{``This [\textbf{$l_q$}] doesn't have a favorite spot. It is equally likely to be on any receptacle it can sit on.''}

\item \textcolor{hab3}{Habit~\#3}: \texttt{``This [\textbf{$l_q$}] moves between receptacles quickly without dwelling. Past placements aren't strong predictors of where it is now. Recently seen at [\textbf{$r_1, r_2, r_3$}].''}

\end{itemize}
where $r_i$ and $p_i$ are filled from the target's empirical placement history in the environment. This grounds the \ac{LLM} estimator in the dataset's residence model in the same way the original paper paper used the simulated resident to elicit human knowledge.

\begin{tcolorbox}[
  colback=orange!4!white, colframe=black!55!white,
  sharp corners, boxrule=0.4pt,
  left=5pt, right=5pt, top=5pt, bottom=5pt,
]
\small
\textcolor{red!65!black}{\textbf{System Prompt:}}
You are an expert assistant that estimates where household objects are likely to be found. Given a scene's receptacle list and the object's previous location, you must output a probability for every receptacle, indicating how likely the object currently rests on it. Probabilities may be unnormalized but must be non-negative.\\[4pt]
\textcolor{blue!65!black}{\textbf{User Prompt:}}
Behavioral context: \textbf{[habit-specific hint, see CEG description above]}.\\
You are an agent that has to find an object that could have moved from its previous location. On \textbf{[$\tau$]} the environment has the following layout: \textbf{[$M_\tau$ receptacle list]}. Also, on \textbf{[$\tau$]}, the \textbf{[$l_q$]} is on \textbf{[current receptacle of $l_q$]}. \textbf{[$\tau_f - \tau$]} later, it is now \textbf{[$\tau_f$]}. Estimate the probability that the \textbf{[$l_q$]} is on each of the receptacles listed above on \textbf{[$\tau_f$]}. Probabilities may be unnormalized but must be non-negative. Reply with one line per receptacle in the exact form: \texttt{<receptacle\_id>\textbackslash t<probability>}. Use every receptacle id from the list, no extra commentary or numbering.
\end{tcolorbox}

\boldparagraph{OSG}
The One-Step Greedy planner of~\cite{10160473} uses the original additive-blend score from which CEG can be seen as a simplified variant. It ranks receptacles according to
$
\frac{\alpha_p}{d_i} + (1-\alpha_p)p_i,
$
where $d_i$ is the navigation distance to receptacle $i$ and $p_i$ is the estimated probability that it contains the target. The coefficient $\alpha_p$ controls the trade-off between cost and reward: $\alpha_p=1$ yields purely distance-based selection, while $\alpha_p=0$ yields purely probability-based selection. We set $\alpha_p=0.5$, following the default used in~\cite{wang2024dynamic}. All remaining components are kept identical to the CEG setting described above. In particular, we use the same two probability estimators, namely the scene-prior table $P(r \mid l_q)$ and the LLM-based variant; the same online visit-fail-zero-renormalize procedure; the same AI2-THOR \texttt{GetShortestPathToPoint} distance computation; the same $0.25m$ denominator guard; and the same per-episode receptacle budget of $10$. The LLM variant also uses the prompt shown in the previous box.

\boldparagraph{HOMER}
We adopt the spatio-temporal object-tracking model of~\cite{pmlr-v205-patel23a}. The model represents each scene as a heterogeneous graph whose nodes correspond to rooms, furniture, and objects, and whose edges encode containment and spatial proximity. Given this graph and the observation history, a Graph Translator Network predicts, for each pickupable object, a probability distribution over its possible receptacles at the next timestep.
We train the network on our training split using the official implementation and training protocol. However, instead of fitting one model per household, we train a single cross-environment model. This setting is aligned with the generalization regime evaluated by \fm: the same checkpoint is applied to all minival environments, without any per-house fine-tuning. At test time, we run the network on the environment scene graph and the observation history up to time $\tau$. We then apply a softmax over destination receptacle nodes and extract the row corresponding to the queried target object. This row defines the predicted receptacle distribution at the future time $\tau_f$. The top-$K=10$ receptacles under this distribution are returned, in ranked order, as navigation goals, preserving the one-shot prediction semantics of the original method.

\boldparagraph{SGM}
We use the Scene Graph Memory (SGM) predictor of~\cite{kurenkov2023modeling}. SGM represents the environment as a hierarchical scene graph, with nodes organized as HOUSE $\to$ ROOM $\to$ FURNITURE $\to$ OBJECT, and applies a HEAT neural edge predictor over this structure. The graph is augmented with temporal node and edge counters, which summarize the observation history up to time $\tau$. Given a queried object, the predictor scores hypothetical edges between that object and each candidate receptacle. These scores are converted to sigmoid probabilities and used to rank receptacles directly. We take the top-$K=10$ ranked receptacles as the navigation-goal sequence. Our implementation follows the published SGM architecture, featurizer, prior, and temporal-counter pipeline. The only departure from the original per-household training setup is that we train a single cross-environment checkpoint. This matches the generalization setting evaluated by \fm: the same model is applied to every minival environment, with no per-house fine-tuning.

\boldparagraph{Naive LLM} This approach consists in a direct prompting baseline that asks the \ac{LLM} to rank the receptacles by likelihood of currently hosting the target. The model and the prompting template are identical to those of the LLMPrior baseline used in the distributional evaluation, with the single difference that the requested top-$T$ is raised from $5$ to $10$ to match the ranked-list size of the other ObjNav baselines. We refer to the distributional appendix for the verbatim system and user prompts.

\subsection{Ablation Studies}
We additionally report a full suite of ablation studies to display how each components of our architecture affect the final results. In particular, we consider the following ablation cases.

A first family probes the role of semantic identity, that is, whether the model benefits from knowing what the entities in the problem are rather than only where they sit. In the \textit{no-query-class} case we drop the class of the query object from the conditioning vector, forcing the model to place an object without knowing its category. In the \textit{no-scene-class} case we remove the per-token class embeddings from the scene map, so that furniture and object tokens are described purely by their bounding boxes, with no semantic label. In the \textit{no-type-emb} case we remove the coarser type embedding that distinguishes furniture tokens from object tokens, testing whether even this binary distinction within the scene carries useful signal.

A second family examines two architectural and encoding choices. In the\textit{ no-scene-encoder} case we remove the Transformer encoder that performs self-attention over the scene tokens before they are consumed by the diffusion backbone, letting the cross-attention layers reason over the raw scene representation directly. In the \textit{linear-bbox-embedding} case we replace our structured three-dimensional bounding-box embedding, which decomposes each box into separate direction and scale components processed by dedicated \acp{MLP}, with a single linear projection, in order to assess whether this structured decomposition is worth its additional complexity.
\begin{table}[t]
\scriptsize
\centering
\begin{tabular}{lcccccc}
\toprule
Method & minFDE$(m)\downarrow$ & Rec@1$\uparrow$ & Density$\uparrow$ & Coverage$\uparrow$ & TV$\downarrow$ & JS$\downarrow$ \\
\midrule
\texttt{FlowMaps} (full) & \best{0.342} & \best{0.942} & \best{0.886} & \best{0.999} & \best{0.829} & \best{0.482} \\
\midrule
no-query-class     & 0.438 & 0.930 & 0.438 & 0.997 & 0.922 & 0.580 \\
no-scene-class     & 0.400 & 0.932 & 0.636 & \second{0.998} & 0.894 & 0.544 \\
no-type-emb        & 0.377 & \second{0.937} & 0.674 & 0.997 & 0.885 & 0.533 \\
no-scene-encoder   & \second{0.372} & 0.934 & \second{0.805} & 0.995 & \second{0.844} & \second{0.498} \\
linear-bbox-emb    & 0.376 & 0.934 & 0.755 & 0.995 & 0.865 & 0.514 \\
\bottomrule
\end{tabular}
\caption{Distributional metrics ablations. Results are averaged over the three habits.}
\label{tab:ab1}
\end{table}

\begin{table}[t]
\scriptsize
\centering
\begin{tabular}{lcccccccc}
\toprule
Method & SR@1$\uparrow$ & SR@5$\uparrow$ & SR@10$\uparrow$ & mSR$\uparrow$ & SPL@1$\uparrow$ & SPL@5$\uparrow$ & SPL@10$\uparrow$ & mSPL$\uparrow$ \\
\midrule
\texttt{FlowMaps} (full) & \best{53.06} & \best{68.14} & \best{69.34} & \best{65.86} & \best{42.52} & \best{47.34} & \best{47.50} & \best{46.66} \\
\midrule
no-query-class     & 19.18 & 47.38 & 50.82 & 43.33 & 15.15 & 24.30 & 24.79 & 23.02 \\
no-scene-class     & 39.02 & 59.02 & 59.51 & 55.52 & 31.31 & 37.83 & 37.90 & 36.90 \\
no-type-emb        & \second{47.21} & 63.61 & 64.59 & 60.79 & \second{38.03} & 43.42 & \second{43.61} & \second{42.65} \\
no-scene-encoder   & \second{47.21} & \second{64.59} & \second{65.25} & \second{61.51} & 37.92 & \second{43.47} & 43.54 & 42.62 \\
linear-bbox-emb    & 47.05 & 61.97 & 63.11 & 59.67 & 37.58 & 42.75 & 42.90 & 42.00 \\
\bottomrule
\end{tabular}
\caption{ObjNav ablations. Results are averaged over the three habits.}
\label{tab:ab2}
\end{table}

The ablations are reported in Table~\ref{tab:ab1} and Table~\ref{tab:ab2}. Overall, the ablations show that semantic information is the dominant factor. Removing the query class causes the largest degradation, especially in ObjNav metrics, indicating that the model needs the target object category to generate placements that are useful for downstream navigation. Removing scene-level class information also substantially hurts both distributional quality and navigation performance, while removing only the coarse object/furniture type embedding has a milder but still consistent effect. The architectural ablations are less severe, but still meaningful: removing the scene encoder or replacing the structured bounding box embedding with a linear projection reduces performance across most metrics. This suggests that both contextual reasoning over scene tokens and the structured geometric encoding contribute to the final model, although their impact is smaller than that of semantic conditioning.

\begin{figure}[t!]
    \centering
    \includegraphics[width=\linewidth]{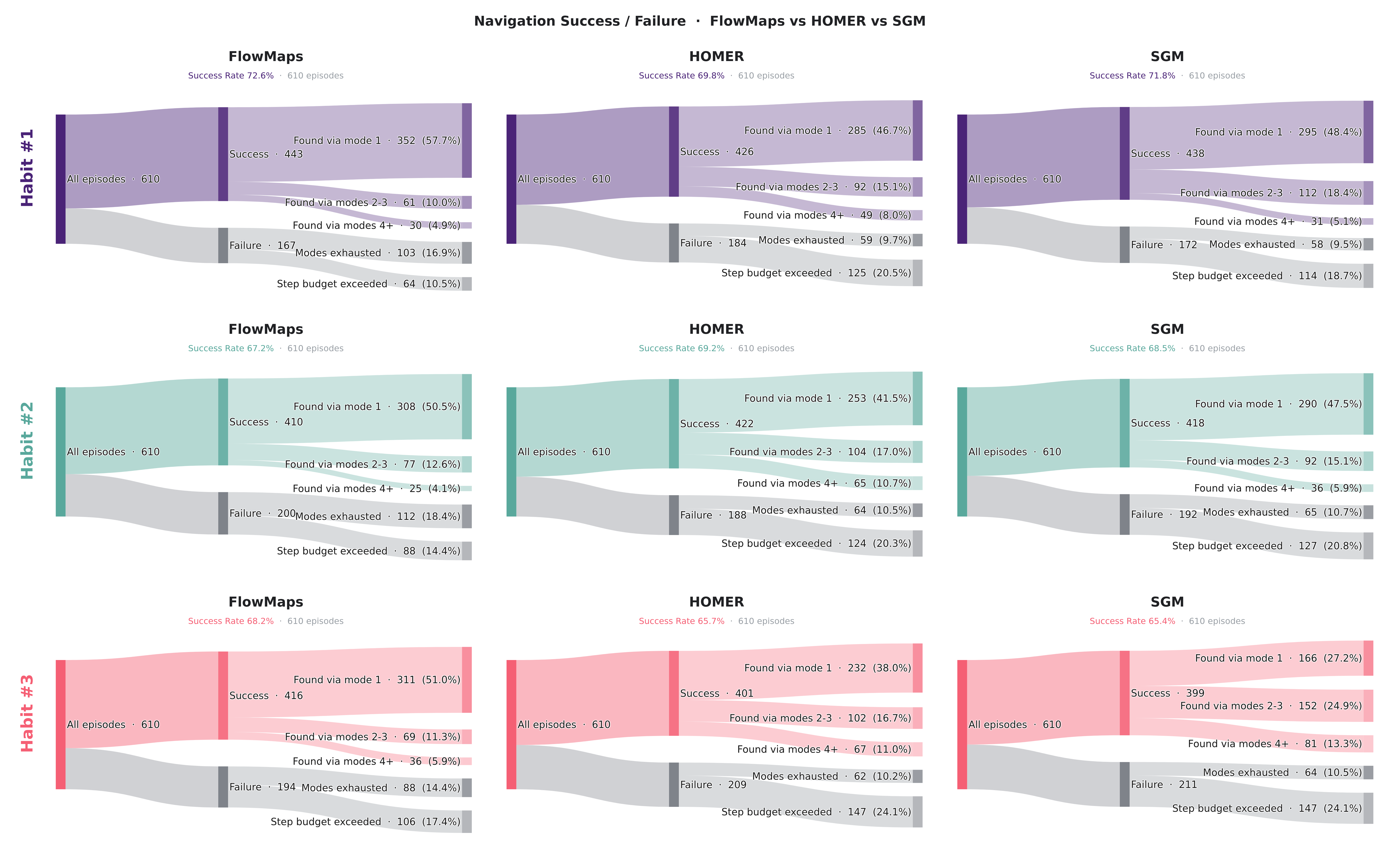}
\caption{
Success and failure decomposition for ObjNav episodes across habits. \fm produces more first-mode successes and fewer step-budget failures, while its remaining failures are mostly due to exhausting the proposed modes.
}
\label{fig:failure_sankey}
\end{figure}

\boldparagraph{Failure analysis} We further analyze how navigation episodes terminate by decomposing each run into successful discoveries at different ranked modes and two failure cases: mode exhaustion, where all proposed targets are visited without observing the object, and step-budget failure, where the agent runs out of steps before completing the ranked search. Fig.~\ref{fig:failure_sankey} reports this decomposition for \fm, HOMER, and SGM across the three habits.

Across all habits, \fm concentrates a larger fraction of successful episodes in the first predicted mode. In \textcolor{hab1}{Habit~\#1}, \fm finds the target at the first mode in 57.7\% of episodes, compared with 46.7\% for HOMER and 48.4\% for SGM. The same trend holds in \textcolor{hab2}{Habit~\#2}, where \fm reaches 49.7\% first-mode success, and in \textcolor{hab3}{Habit~\#3}, where the gap becomes largest: 51.0\% for \fm, compared with 38.0\% for HOMER and 27.2\% for SGM. This indicates that the continuous multimodal distribution learned by \fm is not only useful for covering plausible future locations, but also for ranking them: when the method succeeds, it more often does so without requiring the agent to exhaust later proposals.

The failure modes reveal a complementary trend. \fm consistently produces fewer step-budget failures than the graph-based baselines. For example, in \textcolor{hab1}{Habit~\#1} only 10.5\% of \fm episodes exceed the step budget, compared with 20.5\% for HOMER and 18.7\% for SGM. In \textcolor{hab3}{Habit~\#3}, the same failure accounts for 17.4\% of \fm episodes, versus 24.1\% for both HOMER and SGM. This suggests that \fm proposes targets that are spatially more efficient for the downstream navigation policy, reducing failures caused by long or unproductive search trajectories.
At the same time, \fm exhibits a larger fraction of mode-exhaustion failures. This effect is visible in all three habits, with 16.9\%, 16.4\%, and 14.4\% exhausted episodes for \textcolor{hab1}{Habits~\#1}, \textcolor{hab2}{Habit~\#2}, and \textcolor{hab3}{Habit~\#3}, respectively. By contrast, HOMER and SGM generally have lower exhaustion rates, but compensate with more late-mode successes and more step-budget failures. This points to a distinct error profile: \fm tends to make sharper and more useful early predictions, but when its posterior misses the correct region, additional ranked modes are less likely to recover the object. In contrast, HOMER and SGM spread successful detections over later modes, which can improve recovery in some cases but increases the cost of search. 
Overall, the Sankey analysis shows that \fm' advantage is primarily driven by earlier target localization and fewer navigation-induced failures. The remaining errors are more often due to the predicted distribution missing the correct region than to inefficient traversal. This suggests that future improvements should focus on increasing tail-mode coverage while preserving the strong first-mode ranking behavior, for example by encouraging more diverse samples before clustering or by explicitly calibrating cluster mass during target ranking.
\subsection{Additional Results}

\begin{figure}[h!]
    \centering
    \includegraphics[width=0.9\linewidth]{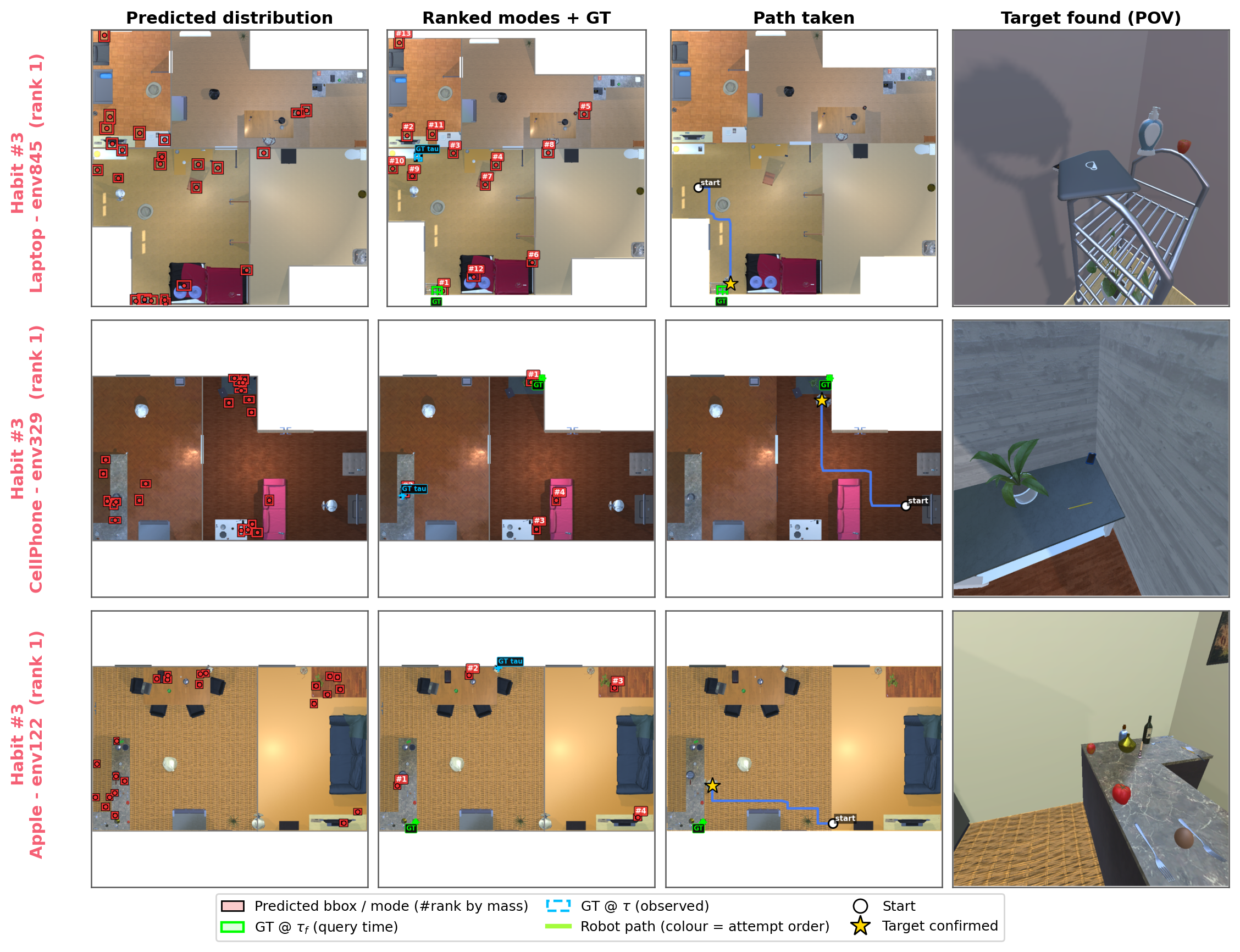}
\caption{Representative successful ObjNav episodes with \fm, set 1. Each row shows the predicted samples, ranked clusters, executed trajectory, and final target confirmation view.}
    \label{fig:p1}
\end{figure}

\begin{figure}[h!]
    \centering
    \includegraphics[width=0.9\linewidth]{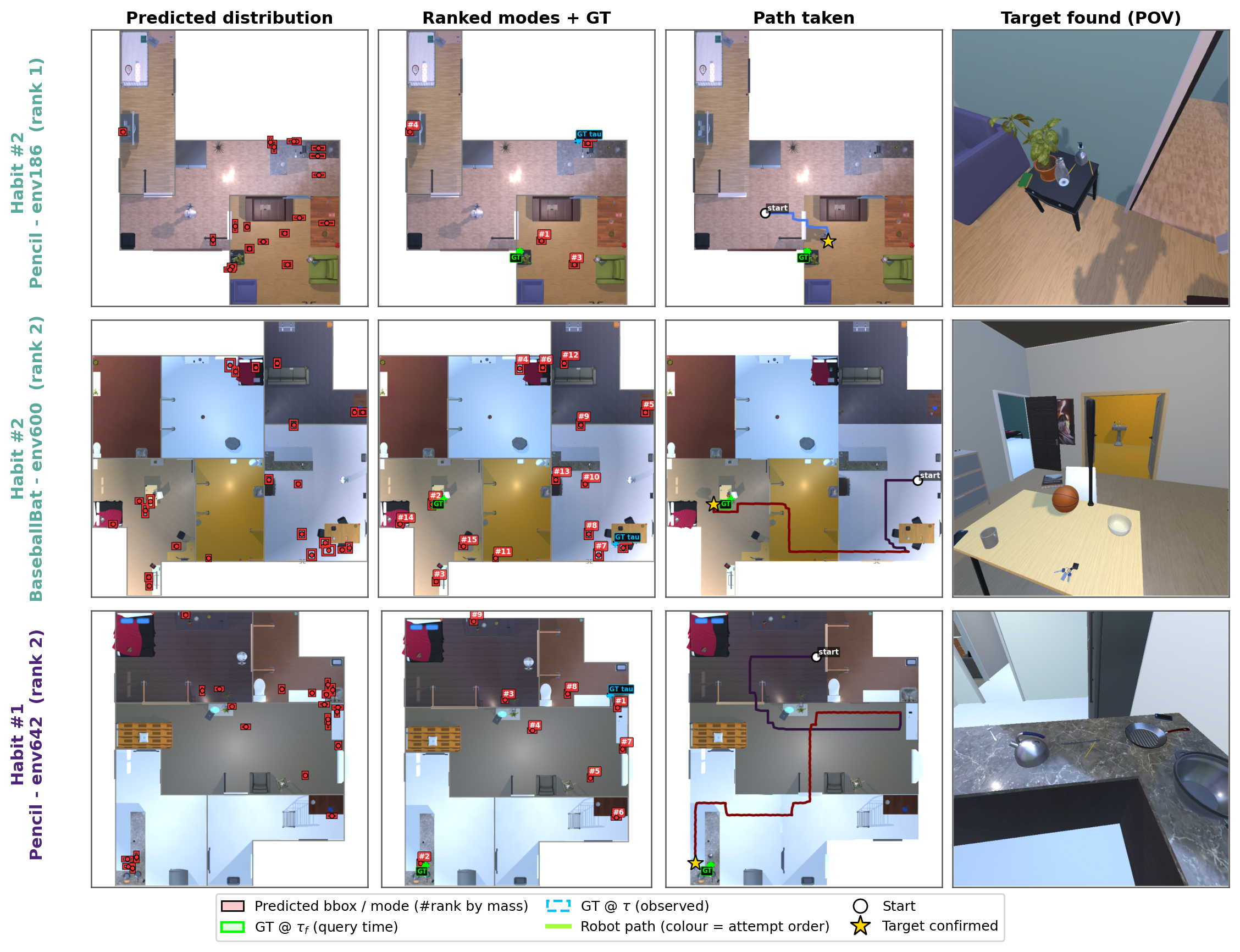}
\caption{Representative successful ObjNav episodes with \fm, set 2. Each row shows the predicted samples, ranked clusters, executed trajectory, and final target confirmation view.}
    \label{fig:p2}
\end{figure}

\begin{figure}[h!]
    \centering
    \includegraphics[width=0.9\linewidth]{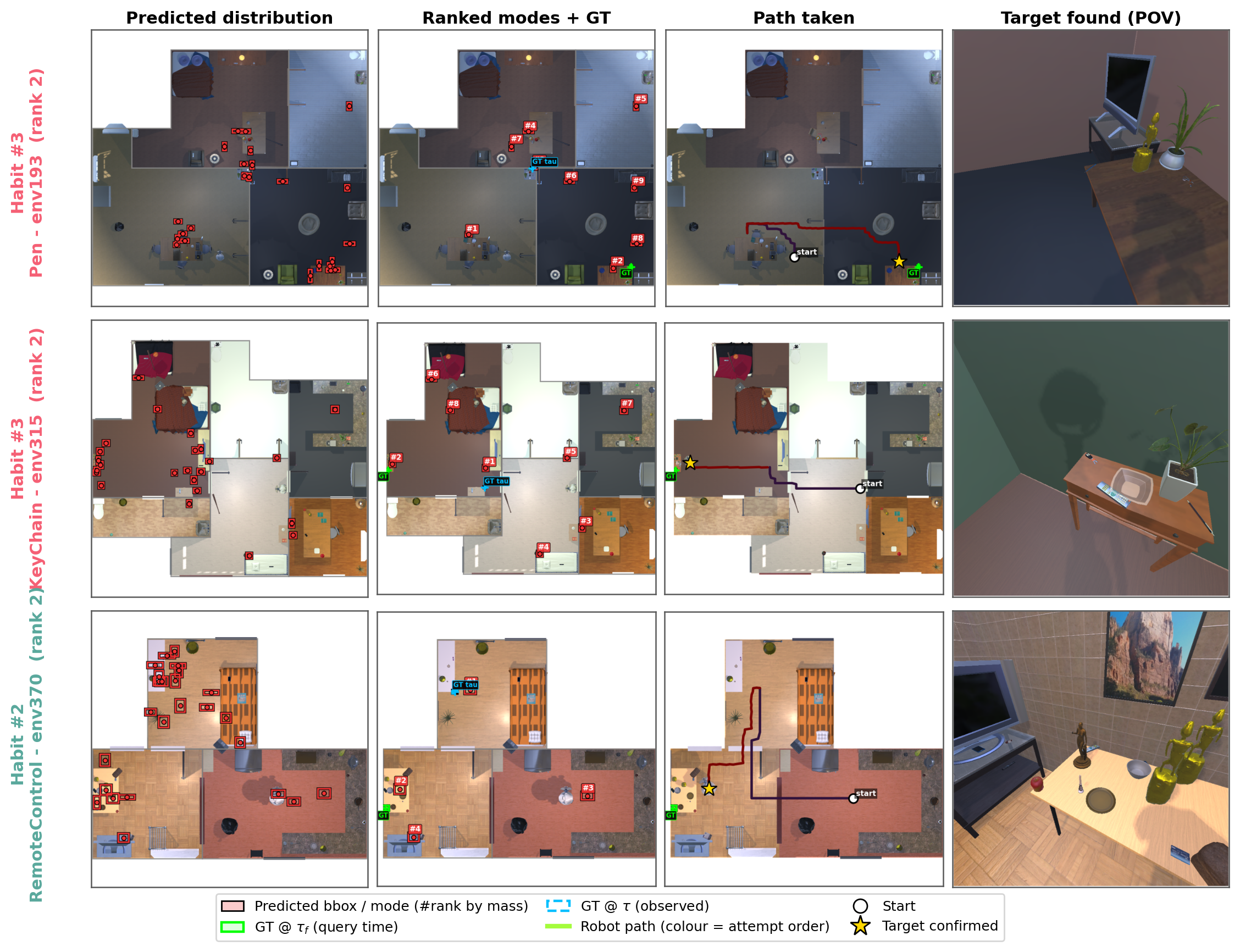}
\caption{Representative successful ObjNav episodes with \fm, set 3. Each row shows the predicted samples, ranked clusters, executed trajectory, and final target confirmation view.}
    \label{fig:p3}
\end{figure}

\begin{figure}[h!]
    \centering
    \includegraphics[width=0.9\linewidth]{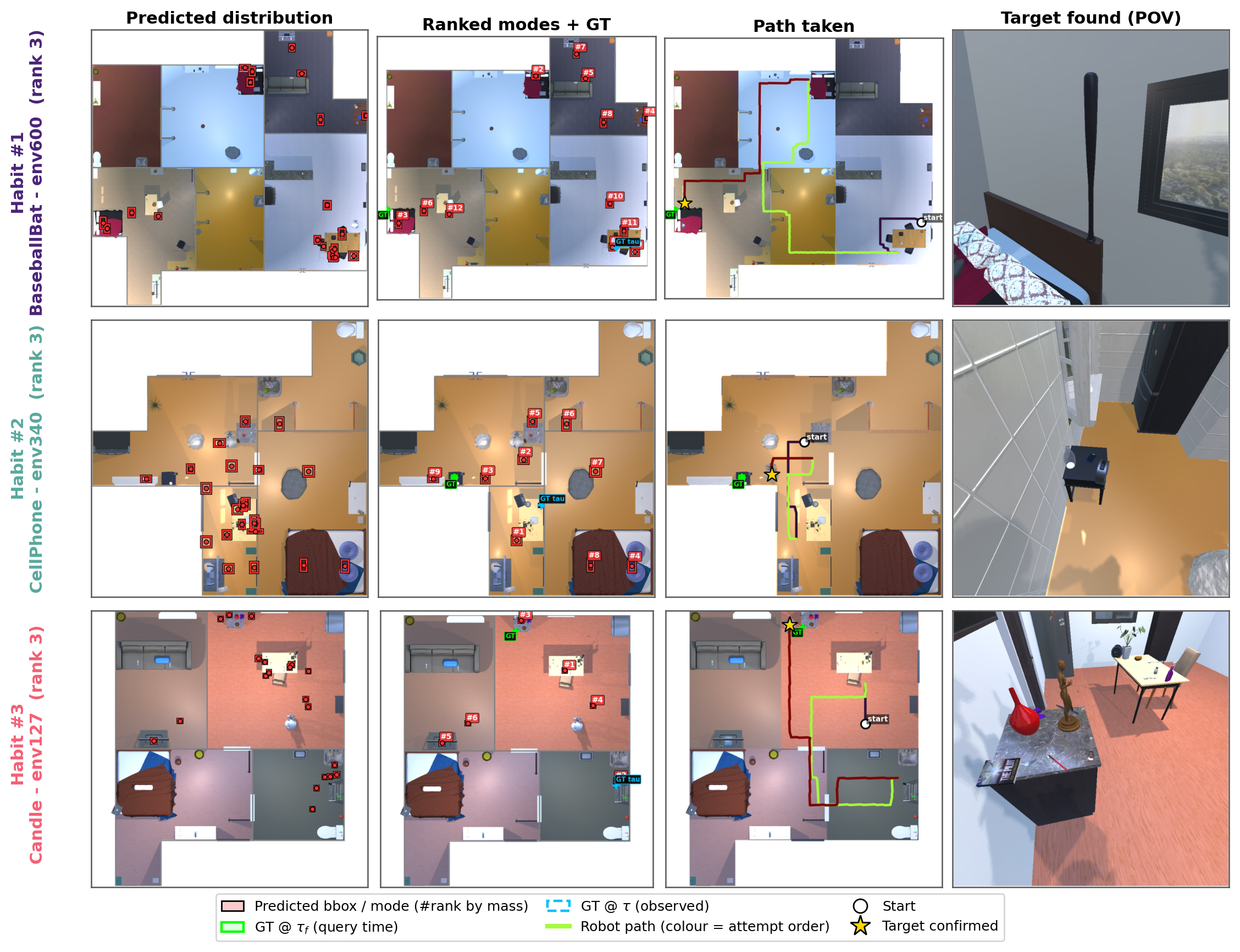}
\caption{Representative successful ObjNav episodes with \fm, set 4. Each row shows the predicted samples, ranked clusters, executed trajectory, and final target confirmation view.}
    \label{fig:p4}

\end{figure}

\begin{figure}[h!]
    \centering
    \includegraphics[width=0.9\linewidth]{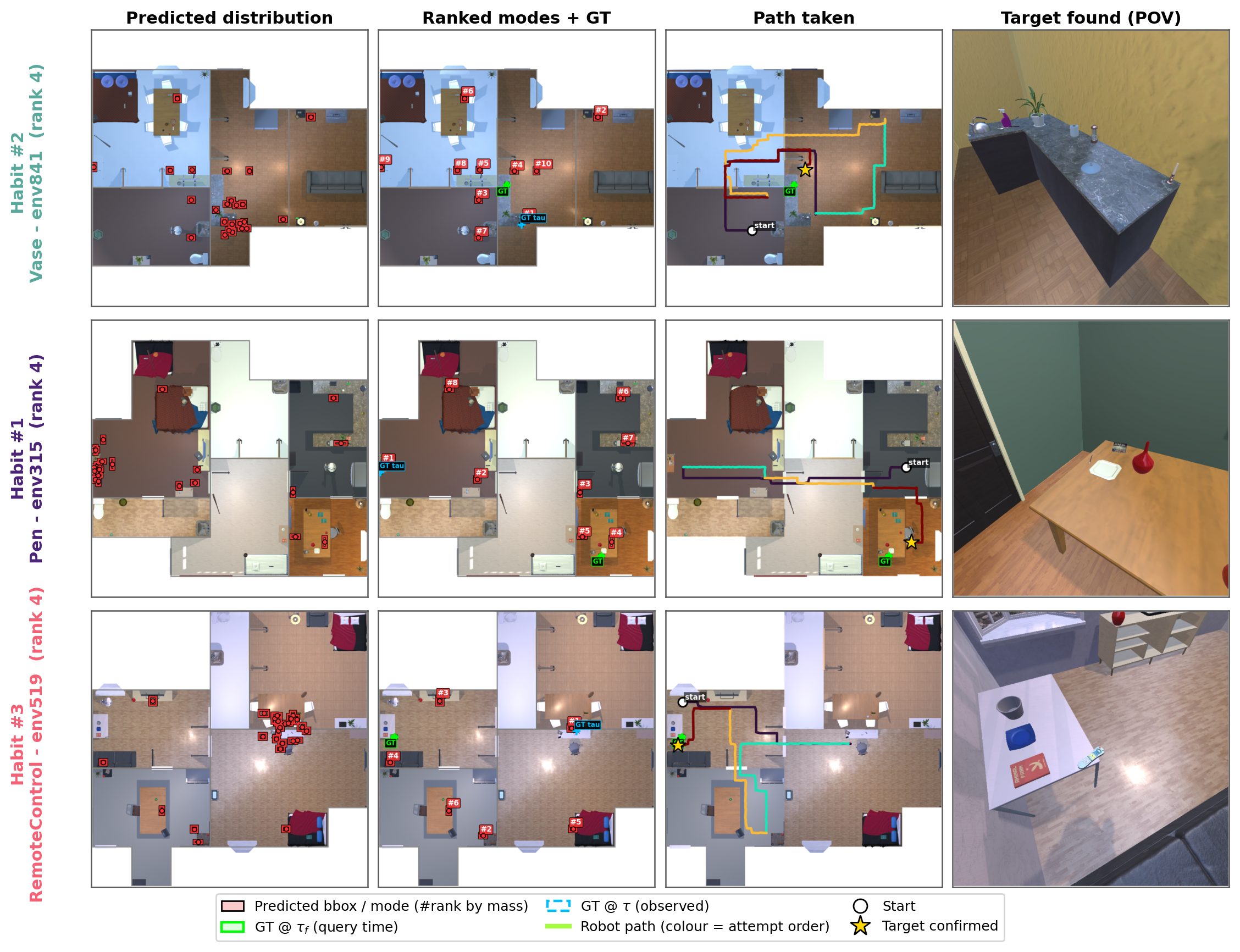}
\caption{Representative successful ObjNav episodes with \fm, set 5. Each row shows the predicted samples, ranked clusters, executed trajectory, and final target confirmation view.}
    \label{fig:p5}

\end{figure}

\begin{figure}[h!]
    \centering
    \includegraphics[width=0.9\linewidth]{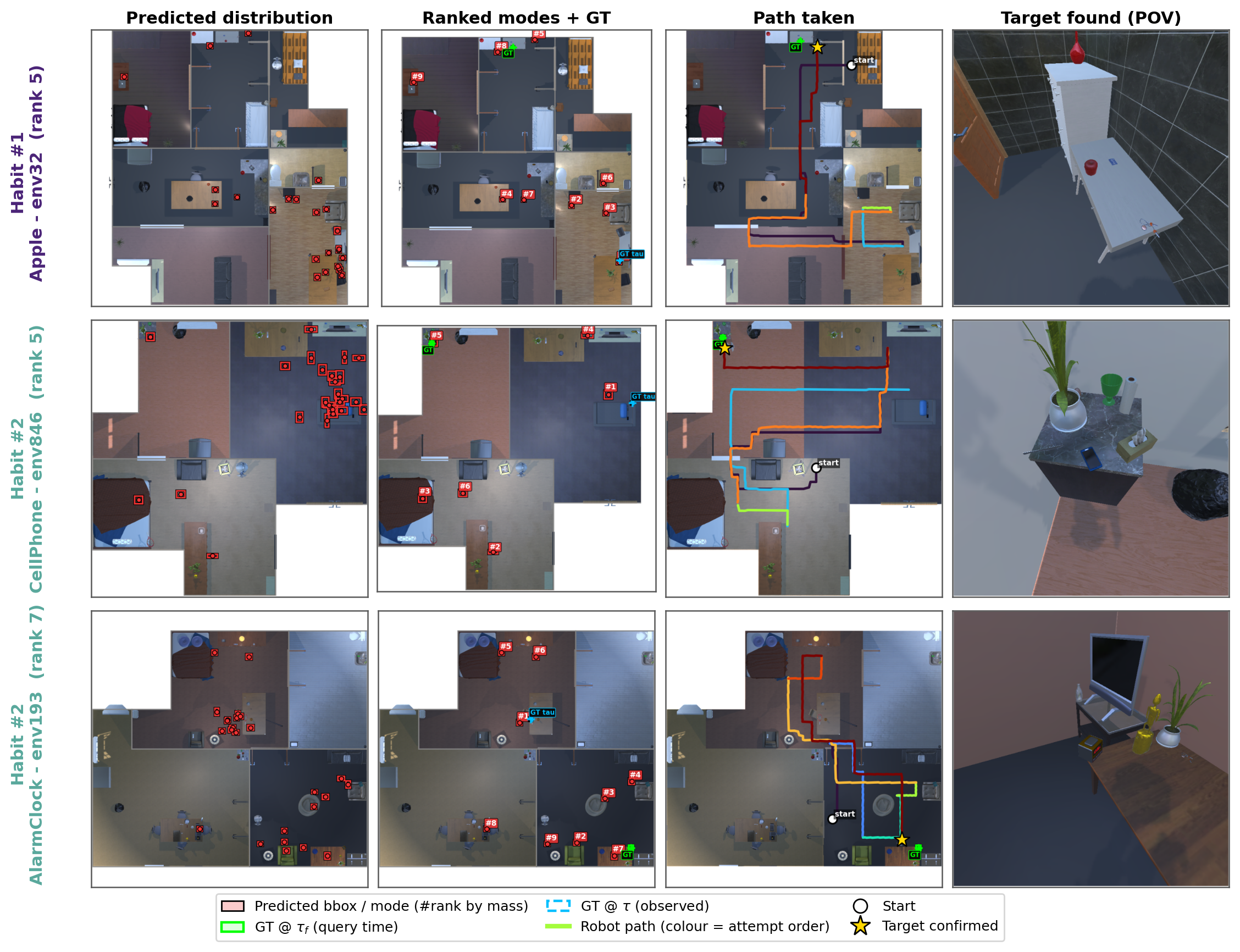}
\caption{Representative successful ObjNav episodes with \fm, set 6. Each row shows the predicted samples, ranked clusters, executed trajectory, and final target confirmation view.}
    \label{fig:p6}

\end{figure}

\begin{figure}[h!]
    \centering
    \includegraphics[width=0.9\linewidth]{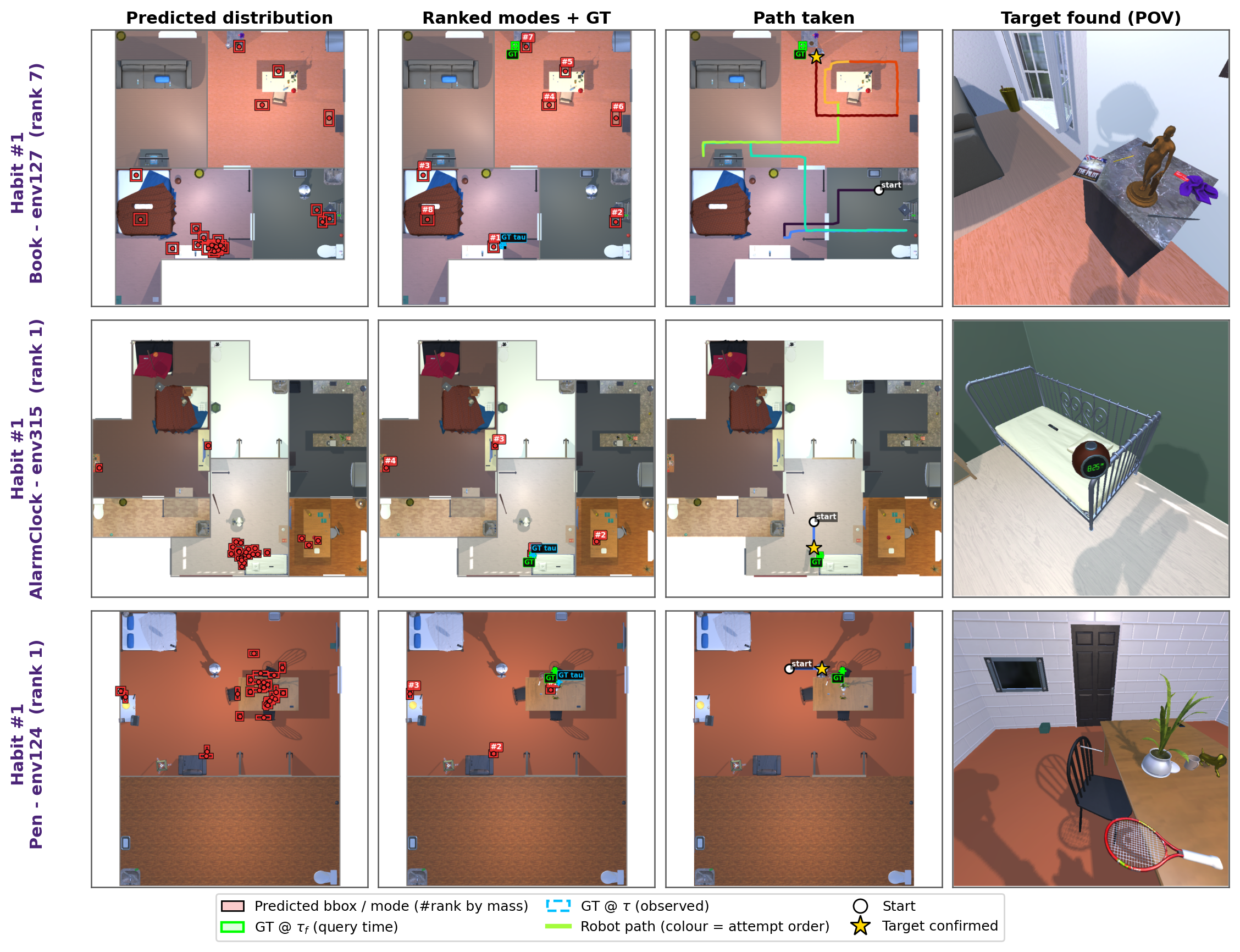}
\caption{Representative successful ObjNav episodes with \fm, set 7. Each row shows the predicted samples, ranked clusters, executed trajectory, and final target confirmation view.}
    \label{fig:p7}

\end{figure}

\begin{figure}[h!]
    \centering
    \includegraphics[width=0.9\linewidth]{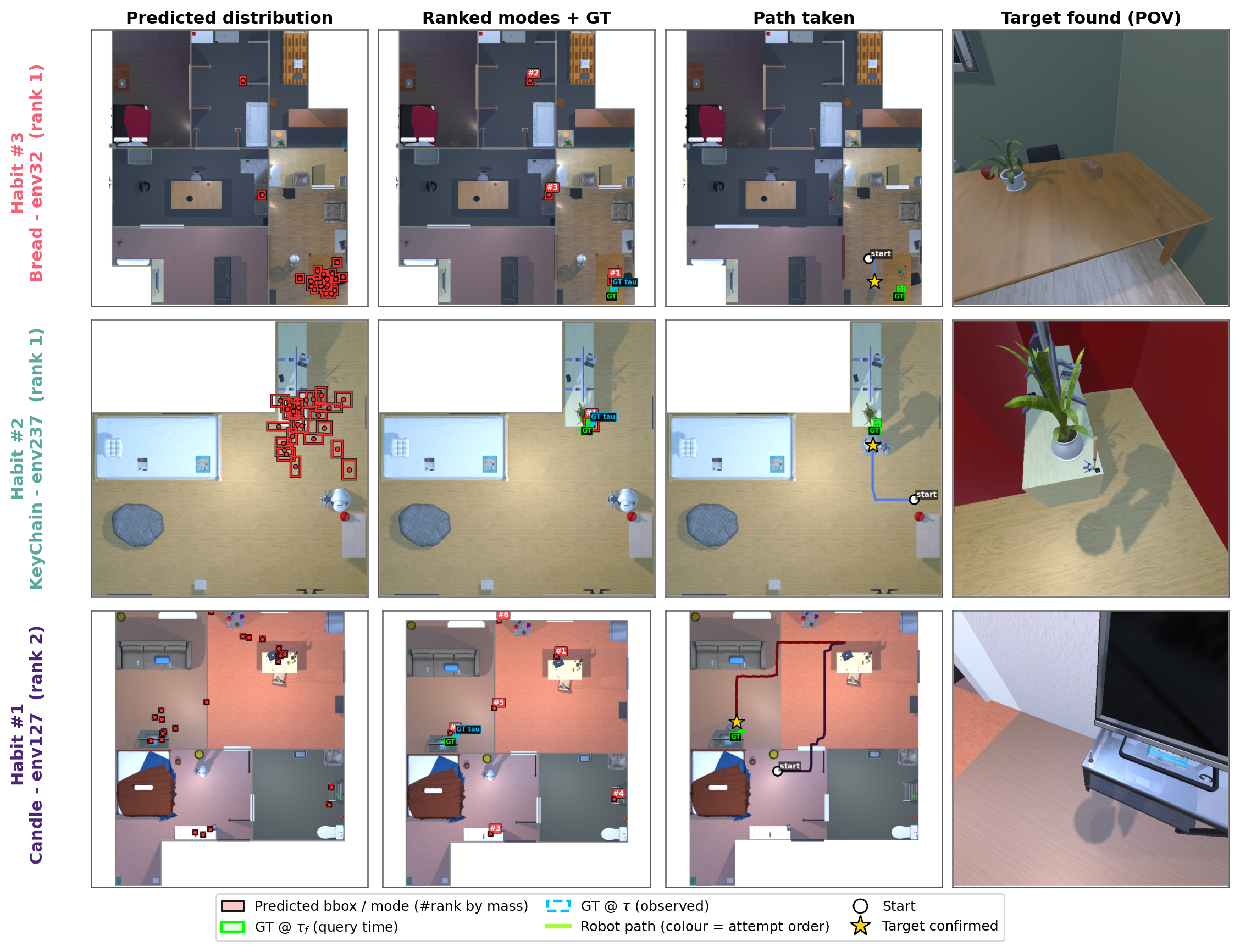}
\caption{Representative successful ObjNav episodes with \fm, set 8. Each row shows the predicted samples, ranked clusters, executed trajectory, and final target confirmation view.}
    \label{fig:p8}

\end{figure}

\boldparagraph{Object Navigation Examples} Figs.~\ref{fig:p1}-\ref{fig:p8} collect representative successful episodes across different habits, environments and objects. Every row reads left to right as one episode. The first panel shows the raw predictive distribution: all $N_{\mathrm{preds}}{=}25$ samples drawn from the flow for the queried object $l_q$ at $\tau_f$, projected onto the top-down map. The second panel overlays the DBSCAN clusters ranked by mass (numbered \#$k$) together with the ground-truth box of $l_q$ both at the observation time $\tau$ (dashed) and at the query time $\tau_f$ (solid): the offset between the two makes explicit that retrieval requires anticipating where the object moved, not recalling where it was last seen. The third panel draws the trajectory the agent actually executed, colored by attempt, as it visits the ranked modes in order from its start pose. The fourth panel is the egocentric view at the moment the target is confirmed. Two patterns recur. For low-depth successes the highest-mass mode lands directly on the $\tau_f$ ground truth and the agent reaches it in a near-straight path, confirming that \fm concentrates probability on the correct future receptacle even when it is far from the last observed location. For higher-depth successes the trajectory visibly re-routes between modes: the agent rules out the leading clusters and falls back through the ranked list, which is exactly the behavior the mass-based ranking is meant to support when the future placement is genuinely multimodal.

%\clearpage
% The acknowledgments are automatically included only in the final and preprint versions of the paper.

%===============================================================================

\clearpage
\bibliography{biblio}  % .bib

\end{document}